\RequirePackage[svgnames]{xcolor}
\documentclass[11pt,letterpaper]{mystyle}

\usepackage[all]{hypcap}
\usepackage[svgnames]{xcolor}
\usepackage[numbers,sort&compress]{natbib}
\usepackage{hyperref}[citecolor=lightblue]

\hypersetup{
    colorlinks = true,
    citecolor = {DodgerBlue},
}
\usepackage{multicol}
\usepackage{setspace}
\usepackage{caption}
\usepackage{ragged2e}
\definecolor{blanchedalmond}{rgb}{1.0, 0.92, 0.8}
\definecolor{carmine}{rgb}{0.59, 0.0, 0.09}
\definecolor{lightblue}{rgb}{0.22,0.45,0.70}%

\renewcommand{\mathbf}{\boldsymbol}

\makeatletter
\def\Ddots{\mathinner{\mkern1mu\raise\p@
\vbox{\kern7\p@\hbox{.}}\mkern2mu
\raise4\p@\hbox{.}\mkern2mu\raise7\p@\hbox{.}\mkern1mu}}
\makeatother

\definecolor{amaranth}{rgb}{0.9, 0.17, 0.31}
\definecolor{antiquebrass}{rgb}{0.8, 0.58, 0.46}
\definecolor{antiquefuchsia}{rgb}{0.57, 0.36, 0.51}
\definecolor{chromeyellow}{rgb}{0.31, 0.47, 0.26}

\newcommand{\github}{\raisebox{-1.5pt}{\includegraphics[height=1.05em]{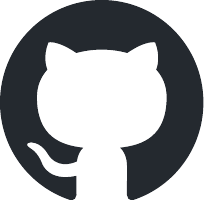}}}

\usepackage{amssymb,mathrsfs,amsmath}

\usepackage[utf8]{inputenc} 
\usepackage[T1]{fontenc}    
\usepackage{url}            
\usepackage{booktabs}       
\usepackage{amsfonts}       
\usepackage{nicefrac}       
\usepackage{microtype}      
\usepackage{array}
\usepackage{graphicx}
\usepackage{wrapfig}
\usepackage{bxcoloremoji}
\usepackage{pifont}
\usepackage{subcaption}
\usepackage{enumitem}
\usepackage[skip=12pt plus 2pt minus 1pt]{parskip}

\usepackage[table]{xcolor}
\usepackage{graphicx}
\usepackage{pifont}
\usepackage{booktabs}
\usepackage{multirow}
\usepackage{makecell}
\usepackage{xcolor}
\usepackage{tabularx}
\usepackage{fontawesome5}

\newcommand{\mbc}[1]{\makebox[2.5em][c]{#1}}

\definecolor{lightgraycustom}{gray}{0.8}
\usepackage{tcolorbox}

\title{MiroEval: Benchmarking Multimodal Deep Research Agents in Process and Outcome}

\author{
  MiroMind Team \\
}

\runningtitle{MiroEval: Benchmarking Multimodal Deep Research Agents in Process and Outcome}

\begin{document}

\begin{abstract}
Recent progress in deep research systems has been impressive, but evaluation still lags behind real user needs. Existing benchmarks predominantly assess final reports using fixed rubrics, failing to evaluate the underlying research process. Most also offer limited multimodal coverage, rely on synthetic tasks that do not reflect real-world query complexity, and cannot be refreshed as knowledge evolves.
To address these gaps, we introduce \textbf{MiroEval}, a benchmark and evaluation framework for deep research systems. The benchmark comprises 100 tasks (70 text-only, 30 multimodal), all grounded in real user needs and constructed via a dual-path pipeline that supports periodic updates, enabling a live and evolving setting.
The proposed evaluation suite assesses deep research systems along three complementary dimensions: adaptive synthesis quality evaluation with task-specific rubrics, agentic factuality verification via active retrieval and reasoning over both web sources and multimodal attachments, and process-centric evaluation audits how the system searches, reasons, and refines throughout its investigation.
Evaluation across 13 systems yields three principal findings: 
the three evaluation dimensions capture complementary aspects of system capability, with each revealing distinct strengths and weaknesses across systems;
process quality serves as a reliable predictor of overall outcome while revealing weaknesses invisible to output-level metrics; 
and multimodal tasks pose substantially greater challenges, with most systems declining by 3 to 10 points. 
The MiroThinker series achieves the most balanced performance, with MiroThinker-H1 ranking the  highest overall in both settings. 
Human verification by three expert annotators confirms benchmark quality at 92.0\% precision. Extensive robustness experiments and a human ranking study (Kendall's $\tau$ = 0.91) further confirm the reliability of the evaluation framework. MiroEval provides a holistic diagnostic tool for the next generation of deep research agents.

 \vspace{4pt}

\coloremojicode{1F4DD} \textbf{Blog Post}: \href{https://miroeval-ai.github.io/blog/}{https://miroeval-ai.github.io/blog/}

\coloremojicode{1F310} \textbf{Project Page}: \href{https://miroeval-ai.github.io/website/}{https://miroeval-ai.github.io/website/}

 \github{} \textbf{GitHub}: \href{https://github.com/MiroMindAI/MiroEval}{https://github.com/MiroMindAI/MiroEval}

\end{abstract}
\maketitle

\begin{figure*}[h]
    \vspace{10pt} 
    \centering
    \includegraphics[width=0.97\linewidth]{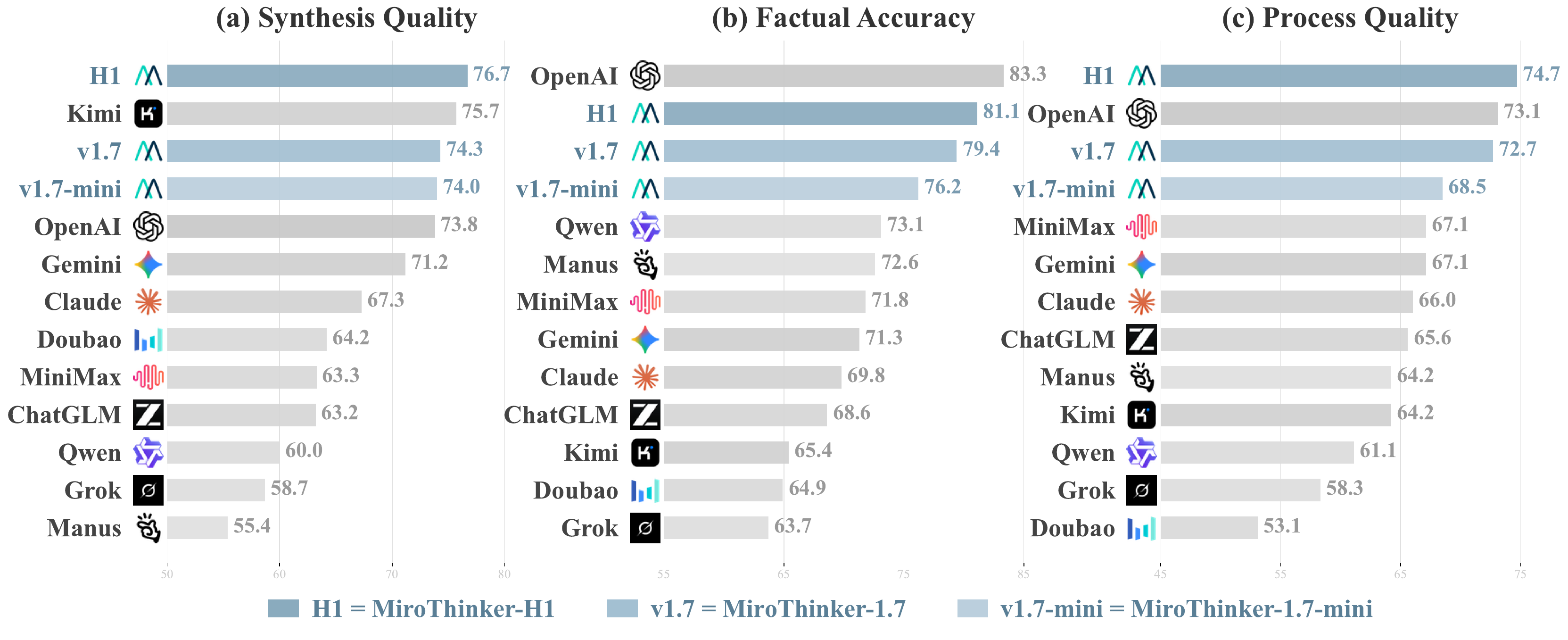}
    \caption{Model performance comparison on 70 text-only deep research tasks across three dimensions.}
    \label{fig:model_performance}
\end{figure*}

\clearpage
\section{Introduction}
The rapid advancement of Large Language Models (LLMs) has driven a pivotal transition from passive text generation to agentic systems capable of autonomous planning and execution~\citep{agentsurvey, li2026just, du2026openseekerdemocratizingfrontiersearch, nguyen2025sfrdeepresearcheffectivereinforcementlearning}.
Deep research, broadly defined as the autonomous, multi-step process of investigating complex information needs through iterative search, evidence gathering, verification, and synthesis~\citep{huang2025deep,zhang2025deep,dong2025doc}, 
has become a prominent agentic paradigm in this transition.
Deep research systems~\citep{kimi2025researcher,manus2025wideresearch,openai2025deepresearch,anthropic2025research,google2025deepresearch} operationalize this paradigm by integrating planning, tool use, heterogeneous source interaction, and long-form report generation into a unified workflow.
As these systems are increasingly adopted in high-stakes domains such as finance, healthcare, and legal analysis, users demand more than a fluent final report: they need answers that are factually reliable, grounded in thorough and traceable investigation, and capable of incorporating multimodal materials (images, PDFs, spreadsheets) that real-world research queries often involve.

Meeting these demands requires continued improvement of deep research systems, which in turn requires reliable ways to measure whether a system truly conducts thorough, factually grounded investigation or merely produces a plausible-looking report. 
Existing benchmarks have made valuable progress in this direction~\citep{abaskohi2025drbench, li2025reportbench, li2026benchmark}, but coverage in several areas remains limited. 
In particular, the majority of existing benchmarks evaluate only the final report, without assessing the underlying research process that produced it~\citep{coelho2025deepresearchgym,li2025reportbench,patel2025deepscholar}.
Multimodal evaluation is rarely supported beyond short-form QA, despite the prevalence of multimodal queries in real-world usage~\citep{li2024survey,huang2026mmdeepresearch,jiang2024mmsearch,foroutan2025wikimixqa}.
Task construction often relies on synthetic or academic queries that do not fully capture the complexity of authentic user needs~\citep{zhu2026gisa,patel2025deepscholar,wang2025liveresearchbench}, and static benchmarks risk becoming stale as the information landscape evolves~\citep{kuissi2026still,thakur2025freshstack}.

To address these challenges, we introduce \textbf{MiroEval}, a benchmark and evaluation framework for deep research systems. The benchmark comprises 100 tasks (70 text-only, 30 multimodal), all grounded in real user needs and constructed through two complementary paths (\S\ref{sec:task_construction}). The first path curates \textbf{65 queries} (35 text-only and 30 multimodal) by rewriting authentic user patterns with privacy-preserving anonymization and difficulty stratification. The second path generates \textbf{35 text-only queries} via an automated pipeline grounded in real-time web trends and validated through a three-stage filtering process to ensure research necessity. Since both paths are driven by analyzable and refreshable data sources, they can be periodically re-executed to keep the benchmark temporally relevant.

The evaluation suite assesses systems through three complementary layers. \textbf{Comprehensive Adaptive Synthesis Quality Evaluation} (\S\ref{sec:comprehensive-eval}) dynamically generates task-specific rubrics and importance weights to assess the final report, moving beyond fixed criteria to capture domain-specific nuances. \textbf{Agentic Factuality Evaluation} (\S\ref{sec:factuality-eval}) decomposes reports into atomic claims and employs an evaluation agent to verify them against both live web sources and multimodal attachments, utilizing a four-way consistency assessment: \texttt{RIGHT}, \texttt{WRONG}, \texttt{CONFLICT}, or \texttt{UNKNOWN}. \textbf{Process-Centric Evaluation} (\S\ref{sec:process-eval}) audits research trajectories across five intrinsic dimensions—search breadth, analytical depth, progressive refinement, critical thinking, and efficiency—while measuring bidirectional alignment between process findings and the final report (Process$\to$Report and Report$\to$Process) alongside contradiction detection, to identify traceability gaps. All three layers natively support multimodal inputs, enabling a holistic diagnostic of the next generation of deep research agents.

Evaluation across 13 leading systems (\S\ref{sec:experiments}) yields three principal findings. First, system rankings shift substantially across synthesis quality, factual precision, and research process rigor, 
demonstrating that each dimension provides non-redundant information. Second, process quality serves as a reliable predictor of overall outcome while also revealing weaknesses invisible to output-level metrics, such as insufficient analytical depth and a significant traceability gap between reports and their underlying research procedures. Third, multimodal tasks pose substantially greater challenges, with most systems declining by 3 to 10 points. 

Further analysis shows that user-derived queries are consistently harder than auto-generated ones while system rankings remain stable across both sources (\S\ref{sec:further-analysis}). Across all dimensions, the MiroThinker series demonstrates the most balanced performance, with MiroThinker-H1 achieving the highest overall scores in both text-only (77.5) and multimodal (74.5) settings. 
Human verification by three expert annotators confirms benchmark quality at 92.0\% precision (\S\ref{subsec:verification}). Extensive robustness experiments and a human ranking study (Kendall's $\tau$ = 0.91) further confirm the reliability of the evaluation framework (\S\ref{sec:further-analysis}). 

\section{Query Collection and Verification}
\label{sec:task_construction}
\begin{figure}[t]
    \centering
    \includegraphics[width=\linewidth]{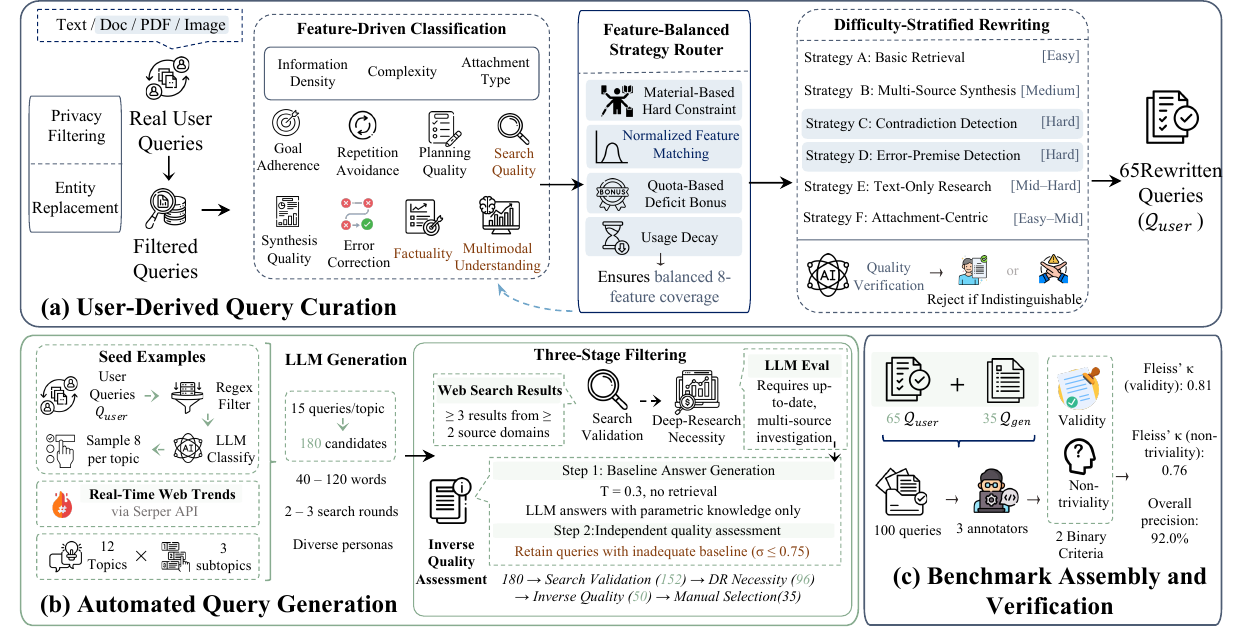}
    \caption{Overview of query construction pipeline.} 
    \label{fig:query_pipeline}
\end{figure}

A reliable benchmark for deep research systems must be grounded in real user needs while maintaining diversity and temporal relevance.
We construct a benchmark of 100 queries via two complementary paths (Figure~\ref{fig:query_pipeline}):
(1)~curating 65 queries (35 text-only and 30 multimodal) inspired by real user query patterns obtained during a closed internal testing phase, with privacy-preserving rewriting and difficulty stratification (\S\ref{subsec:user_derived}); and
(2)~generating 35 text-only queries via an automated pipeline grounded in real-time web trends (\S\ref{subsec:auto_gen}).
We first present an overview of the resulting benchmark (\S\ref{subsec:overview}), then describe each construction path, and finally report quality verification results (\S\ref{subsec:verification}).

\subsection{Benchmark Overview}
\label{subsec:overview}

\begin{figure}[t]
    \centering
    \includegraphics[width=\linewidth]{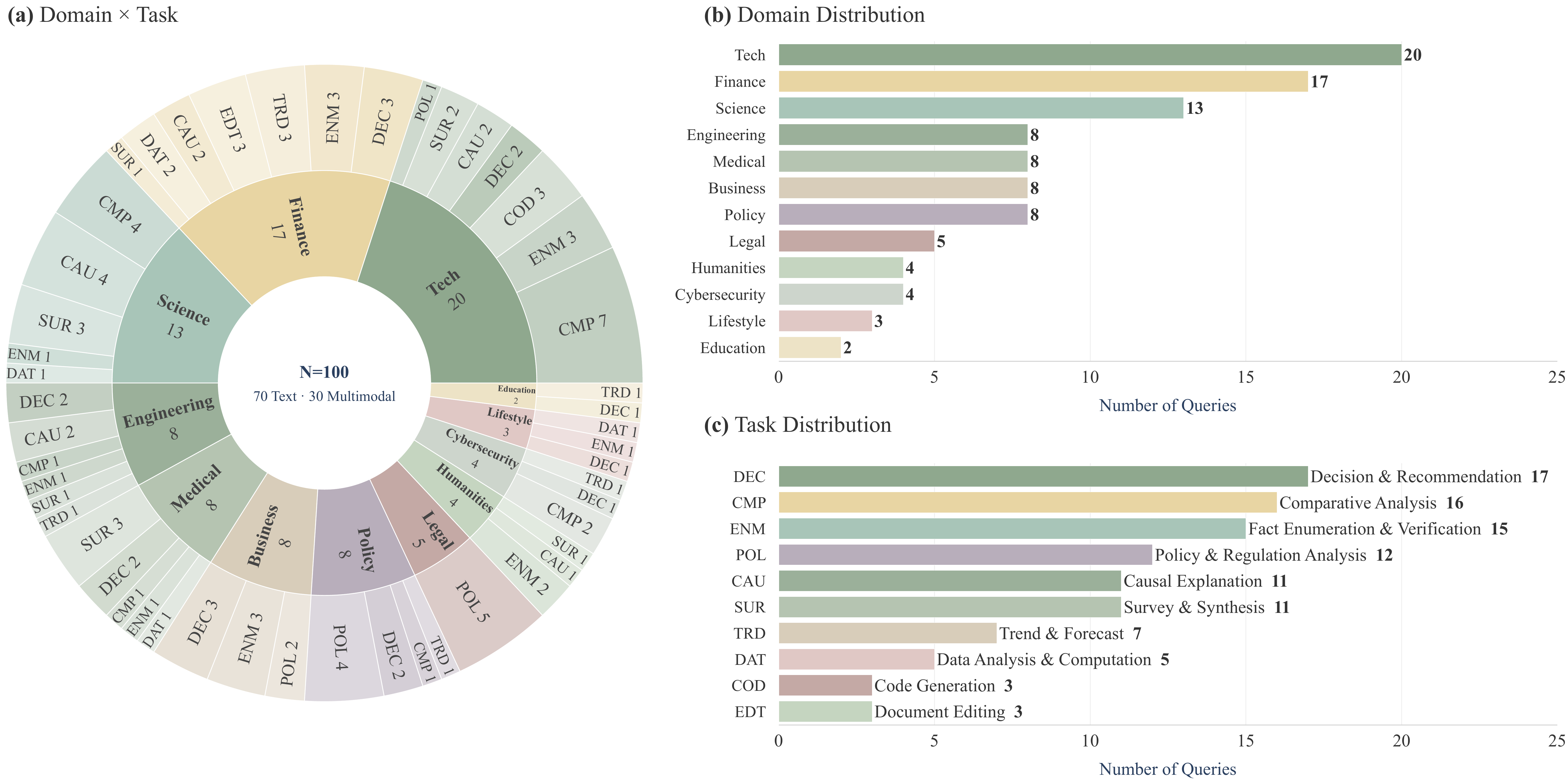}
    \caption{Overview of the query distribution. (a)~Joint distribution of 12 domains and 10 task types. (b)~Domain distribution. (c)~Task type distribution. N\,=\,100 (70 text-only, 30 multimodal).}
    \label{fig:query_distribution}
\end{figure}

The final benchmark comprises 100 queries: 70 text-only and 30 multimodal (Figure~\ref{fig:query_distribution}).

\paragraph{Domain Coverage.}
Queries span 12 domains reflecting the breadth of real-world deep research needs.
Technology (20) and Finance (17) are the most represented, followed by Science (13).
Eight mid-frequency domains (Engineering, Medical, Business, Policy, Legal, Humanities, Cybersecurity, and Education) each contribute 2 to 8 queries, ensuring that evaluation is not dominated by a narrow set of topics.

\paragraph{Task Distribution.}
We annotate each query with one of 10 task types that characterize the reasoning pattern required.
The three most common types are Decision \& Recommendation (17), Comparative Analysis (16), and Fact Enumeration \& Verification (15), which together account for nearly half the benchmark.
Policy \& Regulation Analysis (12), Causal Explanation (11), and Survey \& Synthesis (11) form a second tier.
The remaining four types (Trend \& Forecast, Data Analysis \& Computation, Code Generation, and Document Editing) cover specialized research patterns at lower frequency.

\paragraph{Cross-Distribution.}
Figure~\ref{fig:query_distribution}(a) shows the joint distribution of domains and task types.
Task types are spread across domains rather than concentrated within any single one: for example, Comparative Analysis queries appear in Finance, Science, Engineering, and Policy, while Decision \& Recommendation queries span Tech, Medical, Business, and Legal.
This cross-coverage ensures that the benchmark evaluates domain knowledge and reasoning capabilities jointly rather than in isolation.

\subsection{User-Derived Query Curation}
\label{subsec:user_derived}

The first path draws on query patterns observed during a closed internal testing phase of the \textit{MiroMind} deep research system, covering both text-only and multimodal interactions with attachments (images, PDFs, spreadsheets, slides).
Importantly, no original user query appears in the benchmark in any form.
The pipeline analyzes the distribution and structural characteristics of internal testing queries, then produces entirely new benchmark queries that preserve the topic distribution, complexity profile, and modality coverage of the original population while containing no user-identifiable content.

\paragraph{Privacy-Preserving Processing.}
Throughout the entire pipeline, all data handling follows strict confidentiality protocols: raw queries are processed only on access-controlled internal infrastructure, and all LLMs used for filtering, classification, and rewriting are internally deployed instances that do not transmit data to any external service.
At the entry point, an automated filter removes all queries containing privacy-sensitive content (personal communications, confidential documents, and private financial records) based on rule-based and model-assisted filtering.
For retained queries, all named entities (institutions, organizations, and individuals) are replaced with realistic substitutes of the same type and scale 
through an automated anonymization pipeline, 
ensuring that identifiable entities are systematically replaced before entering subsequent stages.

\paragraph{Classification and Routing.}
An LLM classifies each anonymized query along seven dimensions: attachment type, information density, domain, complexity, attachment role, rewrite potential, and a set of target evaluation features drawn from a taxonomy of 8 capabilities (Appendix~\ref{apd:features}): \textit{goal adherence}, \textit{repetition avoidance}, \textit{planning}, \textit{search}, \textit{report generation}, \textit{factuality}, \textit{error correction}, and \textit{multimodal understanding}.
Based on this classification, each query is routed to one of 6 rewriting strategies spanning three difficulty tiers (\textsc{Easy}, \textsc{Medium}, and \textsc{Hard}; Table~\ref{tab:6strategies}).
Routing incorporates four factors:
(1)~\textit{hard constraints} that exclude strategies incompatible with the query's attachment type;
(2)~\textit{feature matching} that scores how well each strategy's target capabilities align with those of the query;
(3)~\textit{quota bonuses} that up-weight strategies covering underrepresented evaluation features; and
(4)~\textit{usage decay} that penalizes frequently selected strategies to maintain diversity.

\paragraph{Difficulty-Stratified Rewriting.}
Each routed query is rewritten into a benchmark-ready instance by an LLM following the selected strategy.
Easy queries require basic retrieval with attachment comprehension; Medium queries involve multi-step reasoning across heterogeneous sources; Hard queries demand contradiction identification or erroneous-premise detection.

The resulting set of 65 queries covers all 8 evaluation features with balanced representation across difficulty tiers.

\subsection{Automated Query Generation}
\label{subsec:auto_gen}

The second path produces 35 text-only queries through a fully automated pipeline that draws on recurring patterns from user query distributions and grounds generation in current web trends, enabling both temporal relevance and on-demand refresh.

\paragraph{Trend-Grounded Generation.}
We organize generation around 12 topics, each with 3 subtopics (Appendix~\ref{apd:topic_taxonomy}).
For each topic, we retrieve recent headlines and snippets via the Serper API as trend context.
An LLM then generates 15 candidate queries per topic, conditioned on the topic description, trend context, and anonymized seed exemplars drawn from a broader pool of real user queries.
Each query is designed to require investigation from multiple distinct angles, draw on diverse source types, and adopt a specific persona (e.g., analyst, engineer, journalist, investor, or graduate student).
This produces an initial pool of 180 candidates.

\paragraph{Three-Stage Filtering.}
We apply three filters to progressively remove unsuitable candidates (Table~\ref{tab:filter_retention}).

\begin{itemize}[leftmargin=*, itemsep=0em, topsep=0.0em]
\item \textbf{\textit{Search validation.}}
Each candidate is submitted to a live web search.
We require $\geq$3 results from $\geq$2 distinct domains, removing queries that are too niche or ambiguous.
This retains 152 queries (84.4\%).

\item \textbf{\textit{Deep-research necessity.}}
An LLM evaluates whether each query demands external sources and further investigation beyond parametric knowledge.
We retain queries with necessity confidence $\geq$0.7, yielding 96 queries (63.2\%).

\item \textbf{\textit{Inverse quality assessment.}}
The most discriminative filter targets a key principle: effective benchmark queries should expose the limitations of parametric knowledge.
We first elicit a baseline answer without search access ($T{=}0.3$) using only parametric knowledge, then assess this baseline in a separate call that produces three signals: a continuous quality score $\sigma \in [0, 1]$, a categorical label $\ell \in \{\texttt{low}, \texttt{medium}, \texttt{high}\}$, and a binary \texttt{requires\_search} flag.
We retain only queries where the baseline is demonstrably inadequate:
\end{itemize}
\begin{equation}
    \mathcal{Q}_{\text{gen}} = \left\{q \;\middle|\; \sigma(q) \leq 0.75 \;\wedge\; \ell(q) \neq \texttt{high} \;\wedge\; \texttt{requires\_search}(q) \right\}.
\end{equation}
The joint condition on all three signals provides robustness against boundary cases where any single indicator may be unreliable.
35 queries are selected from the filtered pool as the final auto-generated set.

\subsection{Quality Verification}
\label{subsec:verification}

\paragraph{Assembly.}
The final benchmark combines 65 user-derived and 35 auto-generated queries for a total of 100 (Table~\ref{tab:filter_retention}).
Each query is annotated with its source, a domain label from 12 categories (Appendix~\ref{apd:topic_taxonomy}), a task type, and source-specific metadata: feature vector and difficulty tier for user-derived queries; topic, necessity confidence, and baseline quality for auto-generated queries.

\begin{table}[t]
\centering
\small
\caption{Benchmark construction statistics. User-derived queries are fully rewritten from patterns observed during internal testing; auto-generated queries are produced by trend-grounded generation with three-stage filtering.}
\label{tab:filter_retention}
\begin{tabular}{lcc}
\toprule
\textbf{Stage} & \textbf{Count} & \textbf{Retention} \\
\midrule
\multicolumn{3}{l}{\textit{User-Derived Path}} \\
\;\; Internal testing patterns $\rightarrow$ rewritten queries & \textbf{65} & --- \\
\midrule
\multicolumn{3}{l}{\textit{Auto-Generated Path}} \\
\;\; Trend-grounded generation        & 180    & --- \\
\;\; + Search validation              & 152    & 84.4\% \\
\;\; + Deep-research necessity        & 96     & 63.2\% \\
\;\; + Inverse quality assessment     & 50     & 52.1\% \\
\;\; + Manual selection               & \textbf{35}     & --- \\
\;\;\;\; \textit{Cumulative from generation} & & \textit{19.4\%} \\
\midrule
\textbf{Final benchmark} & \textbf{100} & --- \\
\bottomrule
\end{tabular}
\end{table}

\paragraph{Human Verification.}
We validate the pipeline on a sample of queries from both sources.
Three annotators with graduate-level research experience independently assess each query on two criteria:
(1)~\textit{validity}, i.e., whether the query constitutes a legitimate deep-research task, and
(2)~\textit{non-triviality}, i.e., whether it requires web search to answer adequately.
As shown in Table~\ref{tab:human_eval}, both sources achieve substantial inter-annotator agreement ($\kappa > 0.74$) and precision above 90\%.

\begin{table}[t]
\centering
\small
\caption{Human verification results. Three annotators assess validity and non-triviality.}
\label{tab:human_eval}
\begin{tabular}{lcc}
\toprule
\textbf{Metric} & \textbf{User-Derived} & \textbf{Auto-Generated} \\
\midrule
Fleiss' $\kappa$ (validity)       & 0.83 & 0.79 \\
Fleiss' $\kappa$ (non-triviality) & 0.78 & 0.74 \\
Majority-vote precision           & 94.0\% & 90.0\% \\
Unanimous agreement               & 86.0\% & 82.0\% \\
\midrule
\multicolumn{3}{l}{\textit{Aggregated}} \\
Fleiss' $\kappa$ (validity)       & \multicolumn{2}{c}{0.81} \\
Fleiss' $\kappa$ (non-triviality) & \multicolumn{2}{c}{0.76} \\
Overall precision                 & \multicolumn{2}{c}{92.0\%} \\
\bottomrule
\end{tabular}
\end{table}

\paragraph{Temporal Refresh.}
Both construction paths support periodic re-execution: the user-derived path can incorporate new rounds of user queries as they become available, while the auto-generated path can be refreshed at any time with the latest web trends. This design prevents the benchmark from becoming stale and reduces the risk of overfitting to known tasks.

\section{Evaluation Methodology}
To provide a rigorous diagnostic of deep research systems, the MiroEval framework departs from traditional static benchmarks by establishing a multi-layered, agentic evaluation pipeline. Recognizing that a high-quality final report is only one facet of a successful investigation, our methodology decouples the research artifact from the underlying investigative procedure. We introduce an adaptive system that dynamically constructs evaluation rubrics tailored to the specific constraints and modalities of each task. This approach allows for a holistic assessment across three critical dimensions: the synthesis quality of the final report, the factual grounding of claims against heterogeneous evidence sources, and the structural integrity of the research trajectory itself.
\begin{figure}[ht!]
    \centering
    \includegraphics[width=\linewidth]{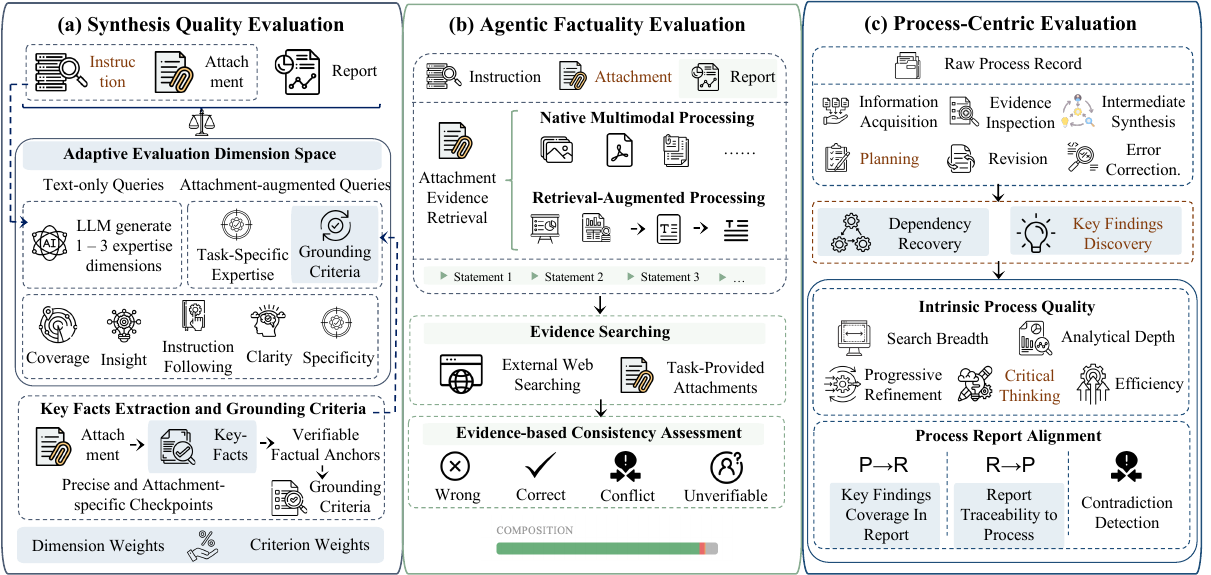}
    \caption{Overview of the evaluation pipeline.
    }
    \label{fig:query_construction}
  \end{figure}

\subsection{Comprehensive Adaptive  Synthesis Quality Evaluation}
\label{sec:comprehensive-eval}

Deep research systems answer complex research queries by performing multi-step retrieval, reasoning, and synthesis to generate long-form, citation-backed reports. Since such tasks vary substantially in domain, objectives, and input modality, fixed evaluation dimension and criteria cannot adequately capture  synthesis quality. To address this challenge, we propose a \emph{Comprehensive Adaptive Synthesis Quality Evaluation} framework that dynamically tailors evaluation dimensions, criteria, and weights to each task.

Queries may involve heterogeneous inputs. In practice, they fall into two categories: (1)~\textbf{text-only queries}, containing only natural-language instructions, and (2)~\textbf{attachment-augmented queries}, where users additionally provide multimodal materials such as images, PDFs, documents, or spreadsheets as supplementary context. The evaluation framework must therefore handle both categories and critically assess whether reports grounded in attachments faithfully leverage the provided materials.

\paragraph{Adaptive Evaluation Dimension Space.}
Let $Q = (I, A)$ denote the input query, where $I$ is the research instruction and $A$ is an optional set of attachments. For each task, the framework constructs a tailored evaluation dimension space
$D = D_{\text{fixed}} \cup D_{\text{dynamic}}(Q).$
The fixed component $D_{\text{fixed}}$
captures universal aspects of synthesis quality, such as Coverage, Insight, Instruction-following, and Clarity. The dynamic component $D_{\text{dynamic}}(Q)$ adapts to the specific characteristics of the query:

\begin{itemize}[nosep,leftmargin=*]
    \item \textbf{Text-only queries} ($A = \emptyset$): the LLM generates 1--3 task-specific expertise dimensions based solely on the instruction $I$ (e.g., ``Policy Pragmatism'' for a cross-national policy comparison).
    \item \textbf{Attachment-augmented queries} ($A \neq \emptyset$): the framework additionally introduces a \textbf{Grounding} dimension, forming composite ``Grounding \& Task-specific Expertise'' dimensions. These dimensions require correct interpretation of attachment content and meaningful analytical expansion, while penalizing superficial referencing or paraphrasing.
\end{itemize}

\paragraph{Key Facts Extraction and Grounding Criteria.}
For attachment-augmented queries, an upstream module extracts condensed \textbf{key facts} from the raw attachments. This process distills heterogeneous materials (e.g., tables from spreadsheets, image captions, and structured text from PDFs or documents) into a set of verifiable factual anchors. These key facts guide the generation of grounding criteria, transforming abstract evaluation requirements into precise and attachment-specific checkpoints.
For example, given a task to ``analyze the global EV market'' with an accompanying sales spreadsheet, a context-free evaluator can only assess general criteria such as whether quantitative analysis is used. However, with extracted key facts, the evaluator can generate concrete criteria such as whether the report correctly identifies the inflection point where BYD surpassed Tesla in 2023Q3. For text-only queries, evaluation criteria are generated directly from the instruction $I$.

\paragraph{Dynamic Weighting and Scoring.}
Given the task-specific dimension space and criteria, the evaluator analyzes $Q$ to derive dimension-level weights $W_d$ and criterion-level weights $w_{d,c}$, subject to constraints $\sum_{d \in D} W_d = 1$ and $\sum_c w_{d,c} = 1$, with explicit justification for each allocation.

The evaluator assesses the report $R$ against each criterion:
\begin{equation}
    s_{d,c} = \text{LLM}_\theta(R,\; d,\; c,\; Q), \quad s_{d,c} \in [0, 10],
\end{equation}
and the final quality score is computed as
\begin{equation}
    S_{\text{quality}} = \sum_{d \in D} W_d \sum_c w_{d,c} \, s_{d,c}.
\end{equation}

\subsection{Agentic Factuality Evaluation}
\label{sec:factuality-eval}

Factuality evaluation assesses whether claims in a generated report are supported by reliable evidence. In deep research scenarios, reports often contain numerous verifiable statements (such as quantities, events, dates, locations, or references to entities) that must be validated against available information sources. Unlike conventional fact-checking settings that assume a single evidence source, real-world research tasks may involve heterogeneous evidence originating from both external web resources and task-provided attachments. These sources may even provide conflicting conclusions, making traditional binary fact-checking insufficient.

Drawing on DeepResearchEval~\citep{wang2026deepresearcheval}, we design an \emph{agentic factuality evaluation} framework based on MiroFlow ~\citep{su2026miroflow} that enables an evaluation agent to retrieve and reason over evidence from multiple sources, following recent advances in long-form factuality verification with agentic or multi-step reasoning \cite{wei2024long,lin2025fact,liu2025verifact}. Given the query $Q=(I, A)$ consisting of research instruction $I$ and attachments $A$ and corresponding report $R$, the system first decomposes it into a set of verifiable statements
\[
\mathcal{S}(Q, R) = \{s_1, \dots, s_n\}.
\]
For each statement $s \in \mathcal{S}(Q, R)$, the agent retrieves supporting or refuting evidence from two complementary sources, forming an evidence set
\[
E(s) = E_{\text{search}}(s) \cup E_{\text{attach}}(s),
\]
where $E_{\text{search}}$ denotes evidence obtained from external search results and $E_{\text{attach}}$ denotes evidence retrieved from task-provided attachments.

\paragraph{Attachment Evidence Retrieval.}
To support factual verification involving uploaded files, the evaluation framework provides a multimodal attachment querying tool that allows the evaluation agent to retrieve evidence from heterogeneous file types. The tool adopts a hybrid processing strategy to accommodate the diverse formats encountered in realistic research scenarios.

\begin{itemize}
\item \textbf{Native Multimodal Processing.}
For file formats that can be directly interpreted by multimodal language models (e.g., images, PDFs, and plain-text documents), the attachment is passed to the model together with the query. The model can then reason directly over visual and structural information such as figures, tables, and document layouts without intermediate conversion.

\item \textbf{Retrieval-Augmented Processing.}
For formats that cannot be directly ingested by the external model (e.g., spreadsheets, slides), the framework applies a retrieval-based approach. The attachment is first converted into textual representations and segmented into smaller chunks. Relevant segments are then retrieved to answer the query, enabling the agent to efficiently locate supporting evidence within large documents.

\end{itemize}

Together, these mechanisms allow the evaluation agent to access and reason over information contained in diverse attachments, enabling the benchmark to evaluate multimodal factual grounding in realistic research scenarios where evidence may originate from both web sources and uploaded files.

\paragraph{Evidence-based Consistency Assessment.}
The agent evaluates the consistency between each statement and its associated evidence set and assigns a factuality label
\[
y(s) \in \{\texttt{RIGHT}, \texttt{WRONG}, \texttt{CONFLICT}, \texttt{UNKNOWN}\}.
\]
The first three labels follow standard fact verification definitions. The additional label \texttt{CONFLICT} captures cases where evidence from different sources leads to inconsistent conclusions, explicitly representing disagreements between heterogeneous information sources rather than forcing them into binary judgments.
\subsection{Process-Centric Evaluation}
\label{sec:process-eval}

While synthesis quality evaluation and factual verification assess the final research artifact, they do not directly evaluate the quality of the underlying research process. In deep research settings, however, process quality is itself an important evaluation target. A system may produce a superficially strong report through redundant exploration or brittle reasoning, while another system may follow a more disciplined and informative process whose intermediate results are only partially reflected in the final write-up. Motivated by this distinction, we introduce a dedicated \emph{process-centric} evaluation framework that focuses on how the system conducts the research procedure, rather than only on the final text it produces. Our framework is organized into three components: process representation, process quality evaluation, and alignment between process-level key findings and report-level key findings.

\paragraph{Process Representation.}
Given a raw process record $P$, we first transform it into a structured process representation that supports downstream analysis. Since raw process logs are often noisy, verbose, and heterogeneous in form, direct evaluation on the original text is unstable and difficult to interpret. We therefore decompose the process into a sequence of atomic units, where each unit corresponds to one functionally distinct step in the research procedure, such as information acquisition, evidence inspection, intermediate synthesis, planning, revision, or error correction. Based on these units, we further recover their local dependency structure and extract the key process findings that emerge during the research procedure. Importantly, this structured representation is used only as an auxiliary analytical interface: its purpose is to make the process more explicit and comparable across tasks and systems, rather than to impose any strong assumption on the exact form of the process itself. 

\paragraph{Process Quality Evaluation.}
Built on the structured representation, we evaluate the intrinsic quality of the research process along several complementary dimensions. 
\begin{itemize}[nosep,leftmargin=*]
\item \textbf{Search Breadth} assesses whether the process explores 
a sufficiently wide range of sources, perspectives, and sub-topics 
relevant to the query.
\item \textbf{Analytical Depth} measures whether the system goes beyond 
surface-level retrieval to conduct multi-step reasoning, follow-up 
investigation, and in-depth analysis of key findings.
\item \textbf{Progressive Refinement} evaluates whether the system 
iteratively improves its understanding over the course of the research, 
refining earlier conclusions as new evidence is gathered.
\item \textbf{Critical Thinking} assesses the system's ability to 
evaluate source reliability, identify limitations in retrieved evidence, 
and respond appropriately to conflicting or weak information.
\item \textbf{Efficiency} measures whether the research process avoids 
unnecessary redundancy, including repeated queries, circular exploration 
paths, and retrieved information that is never utilized.
\end{itemize}
These dimensions are intended to characterize whether the system follows a productive, non-trivial, and self-corrective research process. Unlike report-level evaluation, this component does not directly assess the fluency, stylistic quality, or factual correctness of the final report; instead, it focuses on whether the underlying process exhibits the procedural properties expected from a well-conducted deep research workflow.

\paragraph{Alignment Between Process Findings and Report Findings.}
Beyond intrinsic process quality, we further evaluate whether the final report faithfully reflects the substantive findings developed during the research process. To this end, we extract key findings from the process representation and compare them against the key findings expressed in the final report. This alignment is examined in two directions and one cross-source consistency check.
\begin{itemize}[nosep,leftmargin=*]
\item \textbf{Process$\to$Report (P$\to$R)} checks whether the major findings established during the process are adequately realized in the report. Low P$\to$R scores indicate that useful intermediate results are underutilized or omitted during report synthesis.
\item \textbf{Report$\to$Process (R$\to$P)} checks whether the major conclusions stated in the report can be linked back to sufficient support in the process. Low R$\to$P scores indicate that the report overstates, introduces unsupported synthesis, or departs from the actual research procedure.
\item \textbf{Contradiction Detection (Contr)} evaluates whether the system identifies and resolves conflicting evidence encountered across different sources during research, rather than silently ignoring or propagating inconsistencies into the final report.
\end{itemize}
This component is not intended to duplicate factual verification; rather, it evaluates whether the report is procedurally grounded in the process that produced it, and whether the process itself handles evidentiary conflicts responsibly. Formally, given a process $P$ and final report $R$, the overall process score is defined as
\begin{equation}
S_{\text{process}} = \alpha \, S_{\text{intrinsic}}(P) + (1-\alpha) \, S_{\text{align}}(P, R),
\end{equation}
where $S_{\text{intrinsic}}$ denotes the intrinsic process quality score and $S_{\text{align}}$ denotes the alignment score between process findings and report findings. In this way, the proposed framework complements report-level quality and factual evaluation by explicitly measuring whether the system followed a sound research procedure and whether the final deliverable remains faithful to that procedure.

\section{Evaluation of Deep Research Systems}
\label{sec:experiments}

\subsection{Experiment Setup}

We conduct evaluations on a range of mainstream commercial deep research systems, including OpenAI Deep Research ~\citep{openai2025deepresearch}, Gemini-3.1-Pro Deep Research ~\citep{google2025deepresearch}, Grok Deep Research ~\citep{xai2025grokdeepsearch}, Claude-Opus-4.6 Research ~\citep{anthropic2026claude46}, Manus-1.6-Max Wide Research ~\citep{manus2025wideresearch}, Doubao Deep Research
~\citep{doubao2026doubao}, ChatGLM Agent
~\citep{zhipu2026chatglm}, Kimi-K2.5 Deep Research
~\citep{kimi2025researcher}, Qwen-3.5-Plus Deep Research
~\citep{qwen3.5}, and MiniMax-M2.5 Research 
~\citep{minimax2026m25}. We further include three MiroThinker variants ~\citep{mindteam2026mirothinker}: MiroThinker-1.7-mini, MiroThinker-1.7, and MiroThinker H1. For Kimi-K2.5 Deep Research, Doubao Deep Research, and MiroThinker-1.7-mini, we report only text-only results, as these systems currently do not support multimodal deep research.
For automatic evaluation, we use GPT-5.1 as the judge model for synthesis quality and GPT-5.2 for process evaluation, and GPT-5-mini for factuality evaluation. 

\subsection{Main Results}
\label{sec:main-results}

\newcolumntype{Y}{>{\centering\arraybackslash}X}

\begin{table}[t]
\centering
\footnotesize
\definecolor{rowhl}{RGB}{225,237,237}
\caption{Performance comparison of models with MiroEval.}
\label{tab:main-result}
\renewcommand\arraystretch{1.1}
\setlength{\tabcolsep}{2pt} 

\begin{tabularx}{\linewidth}{@{} l | *{4}{Y} | *{4}{Y} | Y @{}}
\toprule
\multirow{2}{*}{\textbf{Model}} & \multicolumn{4}{c|}{\textbf{Text-Only}} & \multicolumn{4}{c|}{\textbf{MultiModal}} & \multirow{2}{*}{\textbf{Overall}} \\ 
\cmidrule(lr){2-5} \cmidrule(lr){6-9}
& Synthesis & Factuality & Process & Overall & Synthesis & Factuality & Process & Overall & \\ 
\midrule

Kimi-K2.5 Deep Research & 75.7 & 65.4 & 64.2 & 68.4 & -- & -- & -- & -- & -- \\

Doubao Deep Research & 64.2 & 64.9 & 53.1 & 60.7 & -- & -- & -- & -- & -- \\

Grok Deep Research & 58.7 & 63.7 & 58.3 & 60.2 & 56.3 & 71.5 & 53.9 & 60.5 & 60.3 \\

Qwen-3.5-Plus Deep Research & 60.0 & 73.1 & 61.1 & 64.7 & 44.6 & 69.9 & 53.8 & 56.1 & 62.1 \\

Manus-1.6-Max Wide Research & 55.4 & 72.6 & 64.2 & 64.0 & 54.3 & 70.0 & 61.8 & 62.0 & 63.4 \\

ChatGLM Agent & 63.2 & 68.6 & 65.6 & 65.8 & 61.6 & 71.6 & 57.7 & 63.6 & 65.1 \\

MiniMax-M2.5 Research & 63.3 & 71.8 & 67.1 & 67.4 & 56.7 & 71.0 & 62.2 & 63.3 & 66.2 \\

Claude-Opus-4.6 Research & 67.3 & 69.8 & 66.0 & 67.7 & 62.5 & 70.7 & 65.9 & 66.4 & 67.3 \\

Gemini-3.1-Pro Deep Research & 71.2 & 71.3 & 67.1 & 69.9 & 66.4 & 73.7 & 64.1 & 68.1 & 69.3 \\

OpenAI Deep Research & 73.8 & \textbf{83.3} & 73.1 & 76.7 & 66.7 & 77.0 & 66.8 & 70.2 & 74.8 \\

\midrule
\rowcolor{rowhl} MiroThinker-1.7-mini & 74.0 & 76.2 & 68.5 & 72.9 & -- & -- & -- & -- & -- \\

\rowcolor{rowhl} MiroThinker-1.7 & 74.3 & 79.4 & 72.7 & 75.5 & 69.0 & 78.4 & 67.4 & 71.6 & 74.3 \\

\rowcolor{rowhl} MiroThinker-H1 & \textbf{76.7} & 81.1 & \textbf{74.7} & \textbf{77.5} & \textbf{71.5} & \textbf{78.5} & \textbf{73.5} & \textbf{74.5} & \textbf{76.6} \\

\bottomrule
\end{tabularx}
\end{table}

Table~\ref{tab:main-result} presents the performance of all evaluated systems across Synthesis quality, Factuality, and Process under both the Text-Only and MultiModal settings.

\paragraph{Overall Results.}
In the Text-Only setting, systems separate into roughly three performance tiers. MiroThinker-H1, OpenAI Deep Research, and MiroThinker-1.7 form the top tier at 77.5, 76.7, and 75.5 respectively, with MiroThinker-1.7-mini close behind at 72.9. Gemini-3.1-Pro, Kimi-K2.5, MiniMax-M2.5, and ChatGLM Agent constitute a middle tier, spanning approximately 66 to 70. A lower tier includes Manus-1.6-Max , Qwen-3.5-Plus, Claude-Opus-4.6, Doubao and Grok, all scoring below 65, with Grok trailing at 60.2. A broadly similar grouping holds in the MultiModal setting, though overall scores decrease by 3 to 10 points across systems and the inter-system gaps narrow. MiroThinker-H1 achieves the highest MultiModal score at 74.5, followed by MiroThinker-1.7 at 71.6 and OpenAI Deep Research at 70.2, indicating that these systems' advantages generalize robustly beyond text-only tasks.

\paragraph{Key Findings.}
Beyond the overall ranking, three findings emerge from the dimension-level comparison.
First, \textbf{\emph{rankings shift substantially across evaluation dimensions across systems}}. Kimi-K2.5 achieves the highest Synthesis score among non-MiroThinker systems in the Text-Only setting at 75.7, yet its Factuality of 65.4 ranks near the bottom, trailing OpenAI Deep Research by nearly 18 points on this axis. Conversely, Manus-1.6-Max Wide Research obtains the lowest Synthesis score at 55.4, yet its Factuality of 72.6 surpasses several systems with much stronger reports, including Gemini-3.1-Pro and MiniMax-M2.5. These two cases, from opposite ends of the synthesis-quality spectrum, jointly illustrate that a polished report does not guarantee factual grounding, nor does a factually disciplined system necessarily produce well-structured output. We investigate the sub-metric sources of this divergence in \S\ref{sec:outcome-analysis}.
Second, \textbf{\emph{process quality is broadly predictive of outcome quality}}. Across the Text-Only setting, the top three systems on Process (MiroThinker-H1 at 74.7, OpenAI at 73.1, and MiroThinker-1.7 at 72.7) are also the top three on overall outcome, and the weakest process system, Doubao at 53.1, also produces a near-bottom outcome. While a small number of systems deviate from this trend, the overall alignment suggests that process-level evaluation captures a meaningful signal about final output quality. We provide a detailed analysis, including the relationship between process and individual outcome dimensions, in \S\ref{sec:process-analysis}.
Third, \textbf{\emph{multimodal tasks pose substantially greater challenges}}. Overall scores drop by 3 to 10 points for most systems when moving from the Text-Only to the MultiModal setting, with the tier structure broadly preserved but individual systems showing varying degrees of degradation. MiroThinker-H1 proves the most resilient with a decline of only 3.0 points, while Qwen-3.5-Plus suffers the largest drop at 8.6 points. A detailed cross-setting comparison is provided in \S\ref{sec:further-analysis}.

\begin{table*}[t] 
    \renewcommand\arraystretch{0.99}
    \definecolor{rowhl}{RGB}{225,237,237}
    \caption{Combined evaluation on Synthesis quality and Factuality. Report is assessed across five dimensions (Coverage, Insight, Instruction-following, Clarity, and Query Specification). Factuality is measured by the average right ratio (scaled to $[0, 100]$). Overall is the average of Report Avg and Factuality Ratio. Text-Only comprises 70 tasks and Multimodal comprises 30 tasks.}
    \label{tab:outcome-result}
    \setlength{\tabcolsep}{4.5pt}
    \small
        \centering
        \resizebox{\linewidth}{!}{%
      \begin{tabular}{l|ccccc@{\hskip 5pt}c|cccc@{\hskip 5pt}c|c}
        
        \toprule
        \multirow{2}{*}{\textbf{Model}} 
        & \multicolumn{6}{c|}{\textbf{Synthesis}} 
        & \multicolumn{5}{c|}{\textbf{Factuality}} 
        & \multirow{2}{*}{\textbf{Overall}} \\ 
        \cmidrule(lr){2-7}
        \cmidrule(lr){8-12}
        & \mbc{Cov.} & \mbc{Insight} & \mbc{Instr.} & \mbc{Clarity} & \mbc{Spec.} & \mbc{Avg}
         & \mbc{Right} & \mbc{Wrong} & \mbc{Conf.} & \mbc{Unk.} & \mbc{Ratio}
         & \\ 
    \midrule
    \multicolumn{13}{c}{\textbf{Text-Only (70 Tasks)}}  \\
    \midrule

    Grok Deep Research                  & 67.3 & 56.3 & 74.9 & 64.7 & 51.1 & 58.7 & 1924 & 368 & -- & 699 & 63.7 & 61.2 \\
    Manus-1.6-Max Wide Research         & 61.2 & 54.8 & 67.9 & 65.6 & 48.1 & 55.4 & 1972 & 191 & -- & 459 & 72.6 & 64.0 \\
    Doubao Deep Research                & 72.9 & 62.7 & 74.6 & 67.2 & 58.2 & 64.2 & 3890 & 780 & -- & 1393 & 64.9 & 64.6 \\
    ChatGLM Agent                       & 69.9 & 62.8 & 74.5 & 67.5 & 57.1 & 63.2 & 4096 & 580 & -- & 981 & 68.6 & 65.9 \\
    Qwen-3.5-Plus Deep Research         & 64.0 & 64.7 & 69.9 & 67.8 & 52.6 & 60.0 & 1706 & 244 & -- & 380 & 73.1 & 66.5 \\
    MiniMax-M2.5 Research               & 69.8 & 62.7 & 74.2 & 70.6 & 56.7 & 63.3 & 3872 & 486 & -- & 921 & 71.8 & 67.5 \\
    Claude-Opus-4.6 Research            & 73.3 & 72.0 & 73.5 & 71.2 & 61.1 & 67.3 & 2838 & 338 & -- & 910 & 69.8 & 68.6 \\
    Kimi-K2.5 Deep Research             & 80.4 & 79.8 & 78.6 & 76.3 & \textbf{71.7} & 75.7 & 3702 & 595 & -- & 1256 & 65.4 & 70.6 \\
    Gemini-3.1-Pro Deep Research        & 77.4 & 76.6 & 80.0 & 70.1 & 64.9 & 71.2 & 4039 & 526 & -- & 1068 & 71.3 & 71.3 \\
    OpenAI Deep Research                & 78.2 & 74.3 & 81.6 & 77.1 & 69.1 & 73.8 & 3335 & 170 & -- & 496 & \textbf{83.3} & 78.6 \\
    \midrule
    \rowcolor{rowhl} MiroThinker-1.7-mini                & 78.8 & 75.0 & 84.3 & 78.7 & 68.1 & 74.0 & 3397 & 246 & -- & 802 & 76.2 & 75.1 \\
    \rowcolor{rowhl} MiroThinker-1.7                     & 79.2 & 74.7 & 84.7 & 80.1 & 68.4 & 74.3 & 3334 & 181 & -- & 670 & 79.4 & 76.9 \\
    \rowcolor{rowhl} MiroThinker-H1                      & \textbf{80.6} & \textbf{80.3} & \textbf{84.7} & \textbf{81.0} & 70.0 & \textbf{76.7} & 3746 & 161 & -- & 673 & 81.1 & \textbf{78.9} \\

    \midrule
    \multicolumn{13}{c}{\textbf{Multimodal (30 Tasks)}}  \\
    \midrule

    Qwen-3.5-Plus Deep Research         & 46.8 & 46.3 & 52.9 & 52.6 & 30.1 & 44.6 & 576 & 99 & 19 & 101 & 69.9 & 57.3 \\
    Manus-1.6-Max Wide Research         & 58.7 & 50.2 & 65.0 & 61.2 & 40.4 & 54.3 & 681 & 81 & 32 & 134 & 70.0 & 62.2 \\
    MiniMax-M2.5 Research               & 63.1 & 53.3 & 69.1 & 62.0 & 39.2 & 56.7 & 1255 & 184 & 59 & 255 & 71.0 & 63.8 \\
    Grok Deep Research                  & 61.8 & 52.5 & 68.9 & 60.4 & 40.5 & 56.3 & 734 & 104 & 37 & 163 & 71.5 & 63.9 \\
    ChatGLM Agent                       & 67.1 & 60.2 & 71.7 & 65.4 & 45.1 & 61.6 & 1038 & 144 & 46 & 215 & 71.6 & 66.6 \\
    Claude-Opus-4.6 Research            & 68.9 & 66.8 & 62.8 & 59.3 & 50.0 & 62.5 & 964 & 84 & 44 & 243 & 70.7 & 66.6 \\
    Gemini-3.1-Pro Deep Research        & 72.4 & 70.8 & 72.4 & 62.5 & 50.1 & 66.4 & 1502 & 158 & 94 & 302 & 73.7 & 70.0 \\
    OpenAI Deep Research                & 70.6 & 63.9 & 74.8 & 70.5 & 54.2 & 66.7 & 1062 & 100 & 36 & 157 & 77.0 & 71.8 \\
    \midrule
    \rowcolor{rowhl} MiroThinker-1.7                     & 72.6 & 69.2 & \textbf{78.6} & 75.1 & 53.6 & 69.0 & 1306 & 103 & 63 & 235 & 78.4 & 73.7 \\
    \rowcolor{rowhl} MiroThinker-H1                      & \textbf{72.7} & \textbf{76.0} & 78.6 & \textbf{78.3} & \textbf{59.5} & \textbf{71.5} & 1316 & 82 & 56 & 238 & \textbf{78.5} & \textbf{75.0} \\

        \bottomrule
      \end{tabular}}
    \end{table*}

\paragraph{Consistent Strength of the MiroThinker Series.}
What distinguishes the MiroThinker series from other systems is not dominance on any single dimension, but consistent competitiveness across all three. MiroThinker-H1 achieves the highest overall score in both the Text-Only (77.5) and MultiModal (74.5) settings, ranking first or second on every individual dimension. MiroThinker-1.7 follows closely, ranking among the top three on Synthesis, Factuality, and Process with no significant weakness on any axis. This balanced profile contrasts with other top-performing systems that exhibit clear dimension-specific trade-offs: Kimi-K2.5 excels on Synthesis but lags on Factuality, while OpenAI Deep Research leads on Factuality but is surpassed on Synthesis by multiple systems. Even MiroThinker-1.7-mini, a smaller variant, outperforms the majority of full-scale systems overall. In the following sections, we conduct fine-grained analyses at the outcome level (\S\ref{sec:outcome-analysis}) and the process level (\S\ref{sec:process-analysis}) to investigate the sources of these differences.
\begin{figure}[htbp]
    \centering
    \includegraphics[width=0.95\linewidth]{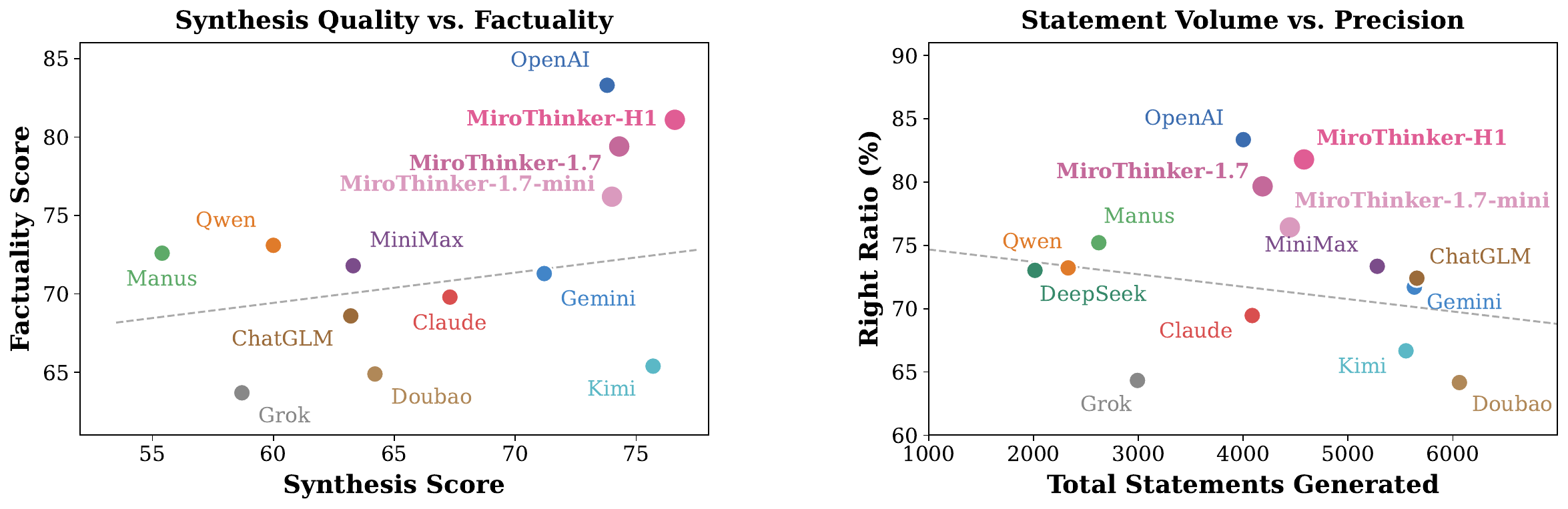}
    \caption{Relationship between synthesis quality, factuality, and statement-level precision across different systems. Left: synthesis quality vs.\ factuality score. Right: total number of generated statements vs.\ right ratio. Each point represents a system. The gray dashed lines denote linear regression fits, illustrating a weak positive correlation between synthesis quality and factuality, and a negative correlation between statement volume and precision.}
    \label{fig:report_factual_scatter}
\end{figure}

\subsection{Outcome-Level Analysis}
\label{sec:outcome-analysis}

Having established that Synthesis quality and Factuality are not interchangeable (\S\ref{sec:main-results}), we now examine the sub-metric structure underlying each dimension to understand \emph{where} and \emph{why} systems diverge. Table~\ref{tab:outcome-result} presents the full breakdown. We focus primarily on the Text-Only setting (70 tasks) due to its broader system coverage.

\paragraph{Synthesis Sub-Metrics: Specificity is the Bottleneck, Insight is the Differentiator.}
Among the five Synthesis sub-metrics, Specificity emerges as the universal bottleneck. It is the lowest-scoring sub-metric for nearly every system, trailing Coverage by 10 to 14 points: OpenAI Deep Research scores 78.2 on Coverage but only 69.1 on Specificity, and Manus-1.6-Max Wide Research shows a similar gap of 13.1 points. Even MiroThinker-H1, the strongest system on Synthesis at 76.7, still lags 10.6 points between these two metrics. This consistent shortfall indicates that current systems can identify relevant topics with reasonable breadth, but struggle to provide the granular, evidence-grounded details that distinguish thorough research from surface-level summaries. Instruction-following, by contrast, is uniformly high among top systems and is no longer a meaningful differentiator.
While Specificity marks the shared weakness, Insight is what most separates systems from one another. Scores range from 54.8 for Manus to 80.3 for MiroThinker-H1, a 25-point spread that is substantially wider than Coverage or Instruction-following. This variance reveals that the ability to synthesize non-obvious analytical observations, rather than merely aggregating retrieved information, is the most discriminative report-writing capability. Notably, several systems with moderate overall performance, such as Gemini-3.1-Pro at 76.6 and Claude-Opus-4.6 at 72.0, score relatively well on Insight, suggesting analytical strengths that are offset by weaknesses in other dimensions.

\begin{table*}[t] 
\renewcommand\arraystretch{0.99}
\definecolor{rowhl}{RGB}{225,237,237}
\caption{Process evaluation results. Intrinsic metrics assess the quality of the research process itself across five dimensions (Search Breadth, Analytical Depth, Progressive Refinement, Critical Thinking, and Efficiency). Alignment metrics measure the consistency between the research process and the final report (Findings$\to$Report coverage, Report$\to$Process traceability, and Contradiction detection). Overall is the weighted average of Intrinsic Avg and Alignment Avg.}
\label{tab:process-result}
\setlength{\tabcolsep}{4.5pt}
\small
    \centering
    \resizebox{\linewidth}{!}{%
  \begin{tabular}{l|ccccc@{\hskip 5pt}c|ccc@{\hskip 5pt}c|c}
    
    \toprule
    \multirow{2}{*}{\textbf{Model}} 
    & \multicolumn{6}{c|}{\textbf{Intrinsic}} 
    & \multicolumn{4}{c|}{\textbf{Alignment}} 
    & \multirow{2}{*}{\textbf{Overall}} \\ 
    \cmidrule(lr){2-7}
    \cmidrule(lr){8-11}
     & \mbc{Brdth} & \mbc{Depth} & \mbc{Refin} & \mbc{Critl} & \mbc{Effic} & \mbc{Avg}
     & \mbc{P$\to$R} & \mbc{R$\to$P} & \mbc{Contr} & \mbc{Avg}
     & \\ 
    \midrule
    \multicolumn{12}{c}{\textbf{Text-Only (70 Tasks)}}  \\
    \midrule

    Doubao Deep Research                & 59.3 & 41.6 & 59.6 & 55.7 & 53.3 & 53.9 & 65.7 & 36.8 & 54.2 & 52.2 & 53.1 \\
    Grok Deep Research                  & 50.9 & 49.4 & 61.0 & 54.6 & 64.7 & 56.1 & 74.6 & 42.2 & 64.6 & 60.4 & 58.3 \\
    Qwen-3.5-Plus Deep Research         & 74.4 & 64.1 & 75.0 & 74.1 & 63.2 & 70.2 & 59.6 & 39.7 & 56.9 & 52.1 & 61.1 \\
    Manus-1.6-Max Wide Research         & 62.8 & 58.4 & 60.6 & 53.5 & 68.8 & 60.8 & 75.1 & 51.3 & 76.3 & 67.6 & 64.2 \\
    Kimi-K2.5 Deep Research             & 77.5 & 59.4 & 71.0 & 67.6 & 53.5 & 65.8 & 70.7 & 46.8 & 70.4 & 62.6 & 64.2 \\
    ChatGLM Agent                       & 76.2 & 59.4 & 67.1 & 59.3 & 59.0 & 64.2 & 77.1 & 51.4 & 72.3 & 67.0 & 65.6 \\
    Claude-Opus-4.6 Research            & \textbf{79.1} & 58.8 & 67.2 & 56.7 & 62.2 & 64.8 & 81.0 & 47.1 & 73.5 & 67.2 & 66.0 \\
    Gemini-3.1-Pro Deep Research        & 75.4 & 66.6 & 75.9 & 64.1 & 59.0 & 68.2 & 72.9 & 50.6 & 74.4 & 66.0 & 67.1 \\
    MiniMax-M2.5 Research               & 71.9 & 62.2 & 70.1 & 62.5 & 63.5 & 66.0 & 77.4 & 53.0 & 74.3 & 68.3 & 67.1 \\
    OpenAI Deep Research                & 77.4 & \textbf{67.3} & \textbf{76.7} & \textbf{74.7} & 63.7 & \textbf{72.0} & 83.6 & 59.0 & 79.9 & 74.1 & 73.1 \\
    \midrule
    \rowcolor{rowhl} MiroThinker-1.7-mini                & 75.5 & 56.3 & 71.3 & 70.9 & 59.0 & 66.6 & 79.7 & 56.3 & 75.2 & 70.4 & 68.5 \\
    \rowcolor{rowhl} MiroThinker-1.7                     & 74.4 & 64.4 & 75.7 & 71.6 & 64.6 & 70.1 & 83.7 & 59.4 & 82.5 & 75.2 & 72.7 \\
    \rowcolor{rowhl} MiroThinker-H1                      & 74.9 & 64.9 & 72.2 & 69.1 & \textbf{71.0} & 70.4 & \textbf{87.0} & \textbf{63.3} & \textbf{86.4} & \textbf{78.9} & \textbf{74.7} \\

    \midrule
    \multicolumn{12}{c}{\textbf{Multimodal (30 Tasks)}}  \\
    \midrule

    Qwen-3.5-Plus Deep Research         & 57.0 & 51.3 & 58.7 & 57.7 & 51.3 & 55.2 & 61.7 & 39.3 & 56.3 & 52.4 & 53.8 \\
    Grok Deep Research                  & 41.9 & 44.3 & 52.4 & 42.4 & 59.5 & 48.1 & 72.4 & 41.4 & 65.2 & 59.7 & 53.9 \\
    ChatGLM Agent                       & 52.7 & 52.3 & 55.7 & 44.7 & 54.3 & 51.9 & 73.0 & 47.0 & 70.7 & 63.6 & 57.7 \\
    Manus-1.6-Max Wide Research         & 52.4 & 57.2 & 60.7 & 43.4 & \textbf{65.9} & 55.9 & 74.5 & 54.5 & 74.1 & 67.7 & 61.8 \\
    MiniMax-M2.5 Research               & 51.0 & 59.0 & 65.0 & 43.7 & 63.0 & 56.3 & 77.0 & 52.0 & 75.0 & 68.0 & 62.2 \\
    Gemini-3.1-Pro Deep Research        & 69.7 & \textbf{65.3} & 71.0 & 58.3 & 47.0 & 62.3 & 75.7 & 49.0 & 73.0 & 65.9 & 64.1 \\
    Claude-Opus-4.6 Research            & \textbf{75.2} & 60.7 & 69.6 & 59.3 & 60.0 & 65.0 & 78.9 & 49.3 & 72.6 & 66.9 & 65.9 \\
    OpenAI Deep Research                & 65.5 & 62.1 & \textbf{73.8} & 70.0 & 54.5 & 65.2 & 77.2 & 56.2 & 72.1 & 68.5 & 66.8 \\
    \midrule
    \rowcolor{rowhl} MiroThinker-1.7                     & 65.0 & 57.0 & 72.0 & 63.0 & 57.7 & 62.9 & 80.7 & 58.7 & 76.0 & 71.8 & 67.4 \\
    \rowcolor{rowhl} MiroThinker-H1                      & 68.6 & 63.1 & 73.4 & \textbf{71.0} & 64.1 & \textbf{68.1} & \textbf{86.6} & \textbf{63.4} & \textbf{86.9} & \textbf{79.0} & \textbf{73.5} \\

    \bottomrule
  \end{tabular}}
\end{table*}

\paragraph{Factual Claims: A Precision--Volume Trade-off.}
The Factuality sub-metrics reveal a fundamental tension between how many claims a system generates and how often those claims are correct (Figure~\ref{fig:report_factual_scatter}). At one extreme, ChatGLM Agent and Gemini-3.1-Pro produce over 4,000 correct claims each, but this high volume comes with 580 and 526 wrong claims respectively, plus over 900 unverifiable ones, pulling their Factuality Ratios down to the low 70s. At the other extreme, OpenAI Deep Research generates fewer correct claims at 3,335, but keeps wrong claims to just 170 and unverifiable claims to 496, achieving the highest per-task right Ratio of 83.3. These profiles reflect fundamentally different generation strategies: broad claim coverage at the cost of precision versus selective generation with strict factual discipline.
The MiroThinker series achieves a distinctive balance between these extremes. MiroThinker-H1 produces the highest claim volume among top-tier systems at 3,746 correct claims while maintaining only 161 wrong ones, the lowest absolute error count of any system and a Ratio of 81.1. MiroThinker-1.7 follows a similar pattern with 3,334 correct and just 181 wrong claims, yielding a Ratio of 79.4. Even MiroThinker-1.7-mini maintains this discipline with 3,397 correct and 246 wrong claims. This consistency across model sizes suggests that the factual discipline is architectural rather than solely a product of scale.

\paragraph{Connecting the Two Dimensions: What Drives the Synthesis--Factuality Misalignment?}
The sub-metric breakdowns above help explain the synthesis--factuality misalignment observed in \S\ref{sec:main-results}. Kimi-K2.5, which achieves the highest Synthesis Avg among non-MiroThinker systems yet one of the lowest Factuality Ratios, turns out to combine the leading Insight score of 79.8 with a high wrong-claim count of 595 and the second largest pool of unverifiable claims at 1,256. In other words, Kimi's reports are analytically rich but insufficiently grounded: it generates insightful interpretations that are not always backed by verifiable evidence. Manus-1.6-Max Wide Research presents the mirror image. Its Insight score of 54.8 is the lowest among all systems, dragging its Synthesis Avg down to 55.4, yet it produces only 191 wrong claims across nearly 2,000 correct ones, yielding a competitive Factuality Ratio of 72.6. Manus appears to prioritize factual caution over analytical depth, a defensible strategy for high-stakes tasks but one that limits report usability. These contrasting profiles suggest that the synthesis--factuality gap is not random: it is systematically driven by how systems balance analytical ambition against factual verification.

\paragraph{Takeaway.}
The outcome-level analysis yields two actionable insights. First, improving \emph{specificity} is the most impactful path to better synthesis quality, as coverage and instruction-following are approaching saturation among top systems. Second, the precision--volume trade-off in factual claims is not inherent: the MiroThinker series demonstrates that high claim volume and low error rates can coexist, suggesting that appropriate research and verification strategies can resolve this tension without sacrificing either dimension.
\subsection{Process-Level Analysis}
\label{sec:process-analysis}

We now turn to the process evaluation, which assesses \emph{how} systems conduct research rather than what they produce. Table~\ref{tab:process-result} reports Intrinsic metrics (Search Breadth, Analytical Depth, Progressive Refinement, Critical Thinking, and Efficiency) and Alignment metrics (Findings$\to$Report coverage, Report$\to$Process traceability, and Contradiction detection).

\paragraph{Intrinsic Quality: Systems Search Wide but Fail to Go Deep.}
The Intrinsic sub-metrics reveal a consistent structural imbalance: most systems achieve reasonable Search Breadth but substantially lower Analytical Depth. In the Text-Only setting, Breadth scores cluster between 71 and 77 for most competitive systems, whereas Depth scores spread far more widely, from 41.6 for Doubao to 67.3 for OpenAI Deep Research. This makes Depth the single most discriminative Intrinsic metric, echoing the role that Specificity plays among Report sub-metrics (\S\ref{sec:outcome-analysis}): the ability to go beyond surface-level retrieval and conduct deeper, multi-step analysis is what separates strong research processes from weak ones. Claude-Opus-4.6 offers a particularly instructive case. Its Breadth of 79.1 is the highest among all the systems, but its Depth of 58.8 trails behind by around 8 points, suggesting a search strategy that retrieves broadly but rarely follows up with targeted, iterative investigation.
Beyond Depth, Efficiency is a universal weakness: even the best system on this metric, MiroThinker-H1 at 68.1, scores well below its performance on other Intrinsic dimensions, and most systems fall in the 53 to 64 range. This indicates that current research processes contain substantial redundancy, including repeated queries, circular exploration paths, and retrieved information that is never utilized, pointing to a clear avenue for future optimization.

\paragraph{Alignment: Findings Reach the Report, but Reports Outrun the Process.}
The Alignment metrics expose a revealing asymmetry between two directions of process-report consistency. Findings$\to$Report (F$\to$R) scores are generally high: MiroThinker-H1 leads at 87.0, with OpenAI Deep Research and MiroThinker-1.7 both exceeding 83, and even mid-tier systems such as MiniMax-M2.5 and ChatGLM Agent remaining above 70. This means that information uncovered during the research process is, for the most part, successfully incorporated into the final report.
Report$\to$Process (R$\to$P) tells a different story. Scores are dramatically lower across the board: even the best system, MiroThinker-H1, achieves only 63.3, and most others fall below 55, with Doubao at 36.8 and Qwen-3.5-Plus at 39.7. The gap between F$\to$R and R$\to$P exceeds 23 points for MiroThinker-H1, 24 points for OpenAI, and approaches 30 points for Doubao, revealing that a substantial portion of report content \emph{cannot be traced back to the research process}. Systems routinely introduce claims, interpretations, or synthesized content that do not originate from their documented search and analysis steps. Whether this reflects implicit reasoning, hallucination, or unlogged intermediate steps, the practical implication is the same: current deep research systems exhibit a significant traceability gap that undermines the auditability of their outputs.
Contradiction detection (Contr) further differentiates systems on a complementary axis. MiroThinker-H1 leads decisively at 86.4, followed by MiroThinker-1.7 at 82.5 and OpenAI at 79.9, while Doubao and Qwen-3.5-Plus score below 57, suggesting limited capacity to handle conflicting sources. This spread of over 30 points highlights contradiction resolution as a critical and highly variable capability for complex research tasks where authoritative sources frequently disagree. 

\paragraph{Process as a Predictor of Outcome Quality.}
In \S\ref{sec:main-results} we noted that process quality is broadly aligned with outcome quality. Here we deepen this observation by examining how Process relates to Synthesis and Factuality individually versus jointly. When correlated with Synthesis alone, the relationship is moderate: Doubao achieves a Synthesis of 64.2 despite a Process score of only 53.1, and Qwen-3.5-Plus attains a Process score of 61.1 that substantially outranks its Synthesis of 60.0 relative to peers. The correlation with Factuality alone is similarly imperfect: Kimi-K2.5's Process score of 64.2 would not predict its unusually low Factuality Ratio of 65.4. However, when Synthesis and Factuality are combined into an overall outcome measure, these individual irregularities partially cancel out, and the alignment with Process becomes stronger. This is because a strong research process benefits both dimensions simultaneously, while the idiosyncratic strategies that inflate one dimension at the expense of the other are averaged away. Empirically, we compute the Pearson correlation coefficient between Process and the combined outcome score, obtaining a strong correlation of 0.88. This quantitative result further substantiates our analysis, confirming that process quality serves as a reliable predictor of overall outcome quality.

\paragraph{Takeaway.}
The process-level analysis identifies two systemic weaknesses shared by current deep research systems. First, Analytical Depth and Efficiency are the primary Intrinsic bottlenecks: systems retrieve broadly but rarely investigate deeply, and much of the retrieval effort is wasted. Second, the F$\to$R versus R$\to$P asymmetry reveals a fundamental traceability gap: reports consistently contain more than what the research process can account for. Despite these weaknesses, process quality remains a reliable predictor of overall outcome, validating process-centric evaluation as a meaningful complement to output-level assessment.
\subsection{Further Analysis}
\label{sec:further-analysis}

\newcolumntype{Y}{>{\centering\arraybackslash}X}
\begin{table}[t]
\centering
\footnotesize
\definecolor{rowhl}{RGB}{225,237,237}
\caption{Performance comparison of models on user derived query and auto generation query.}
\label{tab:rewrite_gen}
\renewcommand\arraystretch{1.1}
\setlength{\tabcolsep}{2pt} 
\begin{tabularx}{\linewidth}{@{} l | *{4}{Y} | *{4}{Y} | Y @{}}
\toprule
\multirow{2}{*}{\textbf{Model}} & \multicolumn{4}{c|}{\textbf{User-Derived}} & \multicolumn{4}{c|}{\textbf{Auto-Generation}} & \multirow{2}{*}{\textbf{Overall}} \\ 
\cmidrule(lr){2-5} \cmidrule(lr){6-9}
& Synthesis & Factuality & Process & Overall & Synthesis & Factuality & Process & Overall & \\ 
\midrule
Grok Deep Research                  & 59.8 & 65.6 & 63.4 & 62.9 & 57.8 & 62.0 & 53.4 & 57.7 & 60.3 \\
Doubao Deep Research                & 63.2 & 60.8 & 47.9 & 57.3 & 65.2 & 68.8 & 58.1 & 64.0 & 60.7 \\
Manus-1.6-Max Wide Research         & 57.1 & 67.9 & 64.6 & 63.2 & 53.7 & 77.2 & 63.9 & 64.9 & 64.1 \\
Qwen-3.5-Plus Deep Research         & 57.9 & 70.3 & 58.7 & 62.3 & 62.1 & 75.8 & 63.5 & 67.1 & 64.7 \\
ChatGLM Agent                       & 61.1 & 62.9 & 64.5 & 62.9 & 65.3 & 74.0 & 66.7 & 68.7 & 65.8 \\
MiniMax-M2.5 Research               & 63.5 & 67.5 & 65.4 & 65.5 & 63.1 & 76.0 & 68.9 & 69.3 & 67.4 \\
Claude-Opus-4.6 Research            & 65.9 & 70.1 & 66.3 & 67.4 & 68.7 & 69.6 & 65.7 & 68.0 & 67.7 \\
Kimi-K2.5 Deep Research             & 74.9 & 63.5 & 64.1 & 67.5 & 76.5 & 67.5 & 64.3 & 69.5 & 68.5 \\
Gemini-3.1-Pro Deep Research        & 70.1 & 69.5 & 65.8 & 68.5 & 72.3 & 73.0 & 68.4 & 71.2 & 69.9 \\
OpenAI Deep Research                & 71.4 & \textbf{80.3} & 71.0 & 74.2 & 76.3 & \textbf{86.4} & \textbf{75.1} & \textbf{79.3} & 76.7 \\
\midrule
\rowcolor{rowhl} MiroThinker-1.7-mini                & 72.9 & 73.1 & 68.5 & 71.5 & 75.2 & 79.3 & 68.5 & 74.3 & 72.9 \\
\rowcolor{rowhl} MiroThinker-1.7                     & 73.6 & 78.5 & 71.2 & 74.4 & 75.0 & 80.5 & 74.3 & 76.6 & 75.5 \\
\rowcolor{rowhl} MiroThinker-H1                      & \textbf{75.2} & 78.4 & \textbf{74.3} & \textbf{76.0} & \textbf{78.2} & 83.7 & 75.1 & 79.0 & \textbf{77.5} \\
\bottomrule
\end{tabularx}
\end{table}

We conduct three supplementary analyses to examine whether the findings from \S\ref{sec:main-results}--\S\ref{sec:process-analysis} are robust across task sources, modality settings, and evaluation configurations.

\paragraph{User-Derived vs.\ Auto-Generated Queries.}
The 70 Text-Only tasks comprise two equally sized subsets: 35 user-derived queries curated from real-world usage patterns through privacy-preserving rewriting (\S2.2), and 35 auto-generated queries produced by a trend-grounded pipeline (\S2.3). Table~\ref{tab:rewrite_gen} compares system performance across these two sources.

Auto-generated queries are consistently easier: nearly all systems with complete data score higher on the auto-generated subset, with overall improvements ranging from 0.6 points for Claude-Opus-4.6 to 6.7 points for Doubao Deep Research. This gap likely reflects the greater complexity and ambiguity inherent in queries inspired by real user needs, which often involve underspecified goals, domain-specific jargon, and multi-faceted information requirements that are difficult to replicate through automated generation. Despite this difficulty gap, the relative ranking of systems remains largely stable across the two subsets. OpenAI Deep Research and the MiroThinker series occupy the top positions in both cases, and the lower tier (Doubao, Qwen-3.5-Plus) is also consistent.
Factuality also shows a systematic source effect: the average Factuality score across systems is approximately 4 to 5 points higher on auto-generated queries, suggesting that trend-grounded queries, which are anchored in recent and well-documented web events, are easier to verify than the more niche topics arising from real usage.

These results carry two implications for benchmark design. First, the ranking stability validates that auto-generated queries provide a reasonable proxy for real-world difficulty, supporting the scalability of automated benchmark construction. Second, the consistent difficulty gap highlights that user-derived queries capture a dimension of complexity that automated generation does not fully reproduce, arguing for the inclusion of both sources in a comprehensive benchmark.

\paragraph{Text-Only vs.\ MultiModal Comparison.}
In \S\ref{sec:main-results} we observed that multimodal tasks amplify existing weaknesses. Here we quantify this effect more systematically. Across the eight systems with both Text-Only and MultiModal overall scores, the average overall score drops by 3.1 points. However, the degradation is highly uneven across systems and dimensions.

By dimension, Synthesis quality suffers the largest average decline at approximately 6 points, with Qwen-3.5-Plus experiencing an extreme drop of 15.4 points (from 60.0 to 44.6) and MiniMax-M2.5 declining by 6.6 points. Process scores decrease by an average of roughly 4 points, with ChatGLM showing the sharpest decline of 7.9 points (from 65.6 to 57.7). In contrast, Factuality Ratios remain remarkably stable, dropping by only 0.2 points on average, suggesting that multimodal tasks do not systematically degrade factual precision.

This pattern reinforces a finding from \S\ref{sec:outcome-analysis}: the multimodal bottleneck lies in report generation (particularly specificity and coverage of visual content) and research process quality (particularly analytical depth), not in factual verification. Systems that already struggle with these capabilities in the Text-Only setting experience disproportionate degradation when visual understanding is required. Notably, MiroThinker-H1 shows the smallest overall decline at 3.0 points, suggesting stronger multimodal integration in its research process, while the relative ranking between systems remains broadly consistent across both settings.

\paragraph{Evaluation Robustness.}
To verify that our findings are not artifacts of a particular evaluation configuration, we conduct three robustness checks (detailed in Appendix~\ref{app:human}). First, re-running the primary GPT judge three times on the MultiModal setting yields Overall standard deviations of only 0.3 to 0.6 across systems, with identical rankings in every run. Second, substituting Gemini as an alternative judge on the Text-Only setting inflates absolute scores by 13 to 17 points on Overall, yet the system ranking is perfectly preserved (Kendall's $\tau$ = 1.0). Third, modifying the judge prompt produces Overall shifts of less than 2 points with no rank changes. We further validate against human judgment through a study with 5 expert annotators ranking 10 systems on 5 sampled queries: the top three systems (MiroThinker-H1, OpenAI Deep Research, MiroThinker-1.7) match exactly, and the largest rank shift is only 2 positions. Together, these results confirm that the comparative conclusions in \S\ref{sec:main-results}--\S\ref{sec:process-analysis} are robust across evaluation configurations.

\section{Related Work and Discussion}
\textbf{Deep Research Systems.} Deep research systems have emerged as a distinct paradigm in which language model agents autonomously plan multi-step web investigations, synthesize evidence across heterogeneous sources, and generate structured, citation-grounded reports~\citep{openai2025deepresearch,google2025deepresearch,anthropic2026claude46,kimi2025researcher,manus2025wideresearch}. Several benchmarks evaluate these capabilities from a search or question-answering perspective. General AgentBench~\citep{li2026benchmark} evaluates general-purpose agents on multi-step reasoning and tool use; BrowseComp~\citep{wei2025browsecomp} measures persistent web navigation; HLE~\citep{phan2025humanity} probes expert-level factual knowledge; and other efforts target search breadth~\citep{wong2025widesearch} or grounded page interaction~\citep{deng2023mind2web}. However, these benchmarks assess retrieval accuracy or short-answer correctness, not the quality of synthesized long-form outputs.

\textbf{Report-Level Evaluation.}
Real-world deep research produces reports, not short answers---motivating report-level evaluation. Most existing report benchmarks are \emph{text-only}: DeepResearchBench~\citep{du2025deepresearch} and DRBench~\citep{abaskohi2025drbench} evaluate synthesis quality via human-annotated rubrics; LiveResearchBench~\citep{wang2025liveresearchbench} introduces temporal grounding; ReportBench~\citep{li2025reportbench} verifies factual grounding of cited claims; and ResearcherBench~\citep{xu2025researcherbench} benchmarks multi-step research workflows. Related text-only efforts further enrich this landscape from complementary angles: DeepScholar-Bench~\citep{patel2025deepscholar} studies generative research synthesis in a live setting, DEER~\citep{han2025deer} strengthens expert-level report assessment with broader document-level verification, Personalized Deep Research~\citep{liang2025towards} incorporates authentic user profiles and personalized information needs, and IDRBench~\citep{feng2026idrbench} begins to evaluate interactive deep research behavior beyond static final outputs.

Recent efforts extend to the \emph{multimodal} setting: MM-BrowseComp~\citep{li2025mm} extends BrowseComp to multimodal retrieval but remains a short-form QA task; MMDeepResearch-Bench~\citep{huang2026mmdeepresearch} evaluates multimodal reports but relies on fixed evaluation dimensions. Additional multimodal benchmarks explore adjacent aspects of research-oriented information seeking: Vision-DeepResearch Benchmark~\citep{zeng2026vision} studies joint visual-textual search, MMSearch~\citep{jiang2024mmsearch} benchmarks multimodal search engines in more realistic web environments, and broader evaluation frameworks such as DeepResearchEval~\citep{wang2026deepresearcheval} and DeepFact~\citep{huang2026deepfact} further reflect growing interest in long-form, grounded, and dynamically maintained research evaluation.

Across all of these lines of work, several common limitations persist: evaluation criteria tend to be fixed and task-agnostic, factual verification is often restricted to cited statements or limited evidence scopes, assessment focuses exclusively on the final output without examining the underlying research process, multimodal evaluation rarely goes beyond short-form QA, and benchmark tasks are rarely grounded in real user needs or designed for temporal refresh. MiroEval addresses these limitations along four axes. For evaluation, it introduces adaptive synthesis quality assessment with dynamically generated task-specific rubrics, agentic factuality verification against both web and attachment evidence, and process-centric evaluation that audits how the system searches, reasons, and refines throughout its investigation. All three layers natively support multimodal inputs. For benchmark construction, it grounds all tasks in real user needs through a dual-path pipeline that supports continuous refresh, ensuring that evaluation remains aligned with the evolving complexity of real-world deep research.

\section{Conclusion}
We introduced \textbf{MiroEval}, a benchmark and evaluation framework for deep research systems, comprising \textbf{100 tasks (70 text-only and 30 multimodal)} assessed through three complementary layers: adaptive synthesis quality, agentic factuality, and process-centric evaluation. Our experiments across \textbf{13 leading systems} show that the three dimensions capture complementary aspects of system capability; that process quality reliably predicts overall outcome while revealing weaknesses invisible to output-level metrics; and that multimodal tasks pose substantially greater challenges. Human verification confirms benchmark quality at 92.0\% precision, and extensive robustness experiments together with a human ranking study (Kendall's $\tau$ = 0.91) validate the reliability of the evaluation framework. MiroEval provides a holistic diagnostic tool for the next generation of deep research agents.

\paragraph{Limitations and Future Work.}
Our process evaluation relies on systems exposing their intermediate reasoning traces, which limits applicability to fully closed-source systems that do not provide such access. Additionally, the factuality evaluation currently identifies cross-source conflicts (e.g., between web evidence and user-provided attachments) but does not yet resolve them: the \texttt{CONFLICT} label flags disagreements without determining which source is correct, an important direction for future work. Looking ahead, we plan to leverage the refreshable dual-path construction pipeline to periodically update the benchmark with new queries reflecting evolving user needs and the latest web trends, ensuring that MiroEval remains temporally relevant as a live benchmark.
\label{sec:related_work}

\clearpage
\bibliography{dreval2}

\clearpage
\section*{Contributors}

\vspace{0.5em}

Fangda Ye\textsuperscript{1,2*}, 
Yuxin Hu\textsuperscript{1,2*}, 
Pengxiang Zhu\textsuperscript{1,2*}, 
Yibo Li\textsuperscript{1,2*}, 
Ziqi Jin\textsuperscript{1,3\dag}, 
Yao Xiao\textsuperscript{1\dag}, 
Yibo Wang\textsuperscript{1}

\vspace{-0.5em}

Lei Wang\textsuperscript{1\ddag},
Zhen Zhang\textsuperscript{1\dag},
Lu Wang\textsuperscript{1\dag}, 
Yue Deng\textsuperscript{1}, 
Bin Wang\textsuperscript{1}, 
Yifan Zhang\textsuperscript{1},  
Liangcai Su\textsuperscript{1}, 
Xinyu Wang\textsuperscript{1}, 
He Zhao\textsuperscript{1}, 
Chen Wei\textsuperscript{1}, 
Qiang Ren\textsuperscript{1} 

\vspace{-0.5em}

Bryan Hooi\textsuperscript{2}, 
An Bo\textsuperscript{1,3}, 
Shuicheng Yan\textsuperscript{2}, 
Lidong Bing\textsuperscript{1}

\vspace{0.5em}

\noindent
\textsuperscript{1} MiroMind AI \\
\textsuperscript{2} National University of Singapore \\
\textsuperscript{3} Nanyang Technological University

\vspace{0.5em}

\noindent{\small 
$^{*}$Co-first author \quad 
$^{\dag}$Core contribution \quad 
$^{\ddag}$Project Lead 
}

\clearpage
\appendix

\section{Data Collection and Report Statistics}

Table~\ref{tab:avg_length_detail} summarizes the report length statistics of all evaluated deep research systems.  All reports were collected in March 2026 within a controlled time window to ensure fair comparison across systems. Reports were generated and downloaded from the official interfaces of each system using automated tools.

We report the average length of valid Deep Research outputs produced by each system across all evaluated tasks. For systems that support both text-only and multimodal deep research, we further report length statistics under both settings. A consistent pattern is that text-only reports are generally longer than their multimodal counterparts. Several systems—including MiroThinker-1.7-mini, DeepSeek DeepThink, Kimi-K2.5 Deep Research, and Doubao Deep Research—do not support multimodal deep research. For these systems, only text-only statistics are reported.

\begin{table}[t]
\centering
\footnotesize
\caption{\textbf{Average report length across different systems.}}
\label{tab:avg_length_detail}
\renewcommand\arraystretch{1.1}
\setlength{\tabcolsep}{4pt}

\begin{tabular}{lccc}
\toprule
\textbf{System} & \textbf{Text-Only} & \textbf{Multimodal} & \textbf{Overall} \\
\midrule
OpenAI Deep Research              & 17,669 & 12,751 & 16,194 \\
MiroThinker-H1                   & 20,442 & 11,802 & 17,850 \\
MiroThinker-1.7                  & 21,138 & 11,293 & 18,185 \\
MiroThinker-1.7-mini             & 21,823 & --     & -- \\
Qwen-3.5-Plus Deep Research      & 24,299 & 9,081  & 19,734 \\
Manus-1.6-Max Wide Research      & 10,263 & 5,585  & 8,860 \\
MiniMax-M2.5 Research            & 26,747 & 9,593  & 21,601 \\
Gemini-3.1-Pro Deep Research     & 49,343 & 32,568 & 44,311 \\
Claude-Opus-4.6 Research         & 23,624 & 20,129 & 22,576 \\
ChatGLM Agent                    & 24,386 & 10,313 & 20,164 \\
Kimi-K2.5 Deep Research          & 61,739 & --     & -- \\
Doubao Deep Research             & 43,160 & --     & -- \\
Grok Deep Research               & 7,585  & 4,977  & 6,803 \\
\bottomrule
\end{tabular}
\end{table}

\section{Evaluation Features and Rewrite Strategies}
\label{apd:features}

Table~\ref{tab:8features} defines the 8 evaluation features used to classify and balance the benchmark queries. Each feature corresponds to a core capability of deep research systems. During query curation (\S\ref{subsec:user_derived}), an LLM assigns a subset of these features to each query, and the routing mechanism ensures balanced coverage across the final benchmark.

\begin{table*}[h]
\centering
\small
\caption{Definitions of the 8 evaluation features.}
\label{tab:8features}
\begin{tabular}{p{3.2cm}p{12cm}}
\toprule
\textbf{Feature} & \textbf{Definition} \\
\midrule
Goal adherence & Whether the system maintains focus on all specified goals and constraints throughout a multi-step task without deviating from original objectives or silently dropping sub-tasks. \\
\addlinespace
Repetition avoidance & Whether the system avoids repeating the same information or analysis across different sections of its output when the query contains multiple similar but distinct sub-tasks. \\
\addlinespace
Planning & Whether the system can decompose a complex query into a coherent, logically ordered sequence of execution steps with clear dependencies between stages. \\
\addlinespace
Search & Whether the system can formulate effective search queries and retrieve relevant external information, rather than relying solely on parametric knowledge or the provided attachments. \\
\addlinespace
Report generation & Whether the system can organize retrieval results into a well-structured, logically coherent report (comparison tables, analytical summaries, or recommendation lists) that synthesizes information from multiple sources. \\
\addlinespace
Factuality & Whether factual claims in the system's output are accurate and verifiable against authoritative sources, with proper citation where appropriate. \\
\addlinespace
Error correction & Whether the system can detect errors, contradictions, or problematic premises in the query or attachments and proactively correct them, rather than blindly following flawed instructions. \\
\addlinespace
Multimodal understanding & Whether the system can correctly parse, interpret, and utilize non-textual information in attachments (charts, tables, images, diagrams, structured data), extracting accurate values and understanding spatial/visual relationships. This feature is only assigned to queries with attachments. \\
\bottomrule
\end{tabular}
\end{table*}

Table~\ref{tab:6strategies} describes the 6 rewrite strategies (A through F) spanning three difficulty tiers. Each strategy transforms a raw user query into a benchmark-ready instance targeting specific evaluation features. The routing mechanism selects the optimal strategy for each query based on material constraints, feature matching, quota bonuses, and usage decay (\S\ref{subsec:user_derived}).

\begin{table*}[h]
\centering
\small
\caption{The 6 rewrite strategies used for user-derived query curation. ``Requires attachments'' indicates the strategy is excluded for text-only queries during routing.}
\label{tab:6strategies}
\begin{tabular}{clp{4.5cm}p{7cm}}
\toprule
\textbf{ID} & \textbf{Difficulty} & \textbf{Target Features} & \textbf{Description} \\
\midrule
A & Easy & search, multimodal understanding & Extract 1--2 key points from the attachment, perform one round of retrieval for supplementary context, and generate a concise response. Requires attachments. \\
\addlinespace
B & Medium & planning, search, report generation, factuality, repetition avoidance & Compare attachment data against at least 2 external public sources and produce a structured comparative analysis report. Requires attachments. \\
\addlinespace
C & Hard & factuality, error correction, multimodal understanding & Embed contradictions between the query text and attachment content (e.g., numerical discrepancies, date misalignment). The system must discover the inconsistency through reading the attachment and/or retrieval. Requires high-density attachments. \\
\addlinespace
D & Hard & error correction, goal adherence & Embed false premises or ambiguous expressions in the query. The system should identify the erroneous premise, correct it, and still complete the core task. \\
\addlinespace
E & Medium / Hard & planning, search, report generation, goal adherence, repetition avoidance & Multi-step research query with no attachment dependency. Answers must be synthesized from multiple public sources through iterative retrieval. \\
\addlinespace
F & Easy / Medium & multimodal understanding, report generation & Primary focus on attachment processing: structured extraction, summarization, format conversion, or cross-page synthesis. Retrieval is auxiliary only. Requires attachments. \\
\bottomrule
\end{tabular}
\end{table*}

\section{Topic Taxonomy and Domain Labels}
\label{apd:topic_taxonomy}

Table~\ref{tab:12topics} lists the 12 topics and 36 subtopics used for trend-grounded automated query generation (\S\ref{subsec:auto_gen}). For each topic, web searches are issued per subtopic to collect recent headlines and snippets as trend context for LLM-based query generation.

\begin{table*}[h]
\centering
\small
\caption{Topic taxonomy for automated query generation: 12 topics, each with 3 subtopics.}
\label{tab:12topics}
\begin{tabular}{clp{10cm}}
\toprule
\textbf{\#} & \textbf{Topic} & \textbf{Subtopics} \\
\midrule
1 & AI Policy \& Regulation & EU AI Act implementation; US state AI laws; AI safety frameworks \\
2 & Cybersecurity & Zero-day exploits; Agentic SOC; AI-powered social engineering \\
3 & Finance \& Macro & Central bank policy; Sovereign debt; Infrastructure investment \\
4 & Crypto \& Digital Assets & Stablecoin regulation; DeFi compliance; CBDC adoption \\
5 & Healthcare \& Pharma & Gene therapy trials; GLP-1 market dynamics; FDA regulatory shifts \\
6 & International Trade & Global supply chain restructuring; Free trade agreements impact; Cross-border regulatory harmonization \\
7 & AI Engineering & {LLM benchmarking; Agentic coding tools; Model deployment architecture} \\
8 & Climate \& Energy & Data center sustainability; Carbon pricing; Grid constraints \\
9 & Education \& Workforce & AI in K-12 policy; Workforce reskilling; Immigration \& talent \\
10 & Legal \& Compliance & AI privilege doctrine; GDPR enforcement; Algorithmic discrimination \\
11 & Biotech \& Science & Computational biology; Quantum computing; Open access publishing \\
12 & Supply Chain \& Industrial & Nearshoring trends; Autonomous logistics; Semiconductor supply \\
\bottomrule
\end{tabular}
\end{table*}


Each query in the benchmark is assigned a domain label from the 11 canonical categories listed in Table~\ref{tab:11domains}. A rule-based normalization function maps free-form domain strings to these labels using substring matching and keyword fallbacks, with \texttt{tech} as the default.

\begin{table*}[h]
\centering
\small
\caption{The 11 canonical domain labels.}
\label{tab:11domains}
\begin{tabular}{clp{10.5cm}}
\toprule
\textbf{\#} & \textbf{Label} & \textbf{Scope} \\
\midrule
1 & \texttt{finance} & Financial markets, investment analysis, banking, macroeconomics, corporate earnings, and economic data. \\
2 & \texttt{policy} & Government policy, regulation, governance, and institutional rule-making at local, national, and international levels. \\
3 & \texttt{tech} & Technology, software engineering, AI/ML, hardware, and internet products. Also serves as the default fallback label. \\
4 & \texttt{cybersecurity} & Digital security threats, defenses, vulnerability research, threat intelligence, and security operations. \\
5 & \texttt{health} & Healthcare, medicine, pharmaceuticals, clinical research, medical devices, and public health. \\
6 & \texttt{science} & Natural sciences, academic research, engineering, mathematics, and research infrastructure. \\
7 & \texttt{education} & Education systems, learning, workforce training, talent development, and professional reskilling. \\
8 & \texttt{legal} & Law, legal practice, compliance, data protection enforcement, and algorithmic accountability. \\
9 & \texttt{energy} & Energy systems, climate policy, carbon markets, sustainability, and data center environmental impact. \\
10 & \texttt{trade} & International trade, supply chains, logistics, manufacturing, and cross-border commerce. \\
11 & \texttt{crypto} & Cryptocurrencies, digital assets, decentralized finance, blockchain technology, and CBDCs. \\
\bottomrule
\end{tabular}
\end{table*}

\clearpage
\section{Evaluation Robustness and Human Study}
\label{app:human}

A key concern for any LLM-based evaluation framework is whether the results are sensitive to random variation, the choice of judge model, or minor prompt differences. We address this through three controlled robustness experiments (\S\ref{app:intra-judge}--\S\ref{app:prompt-sensitivity}), followed by a human study that validates consistency with expert judgments (\S\ref{app:human-study}).

\subsection{Intra-Judge Stability}
\label{app:intra-judge}

LLM-based evaluation can exhibit non-trivial variance across runs due to sampling randomness. To quantify this, we re-run the primary judge configuration (GPT series) two additional times on the MultiModal setting (30 tasks) for four systems: OpenAI Deep Research, Gemini-3.1-Pro, MiroThinker-H1, and MiroThinker-1.7. Together with the original run, this yields three independent evaluations per system. Table~\ref{tab:intra-judge} reports the mean and standard deviation across runs.

\begin{table*}[t]
\centering
\footnotesize
\definecolor{rowhl}{RGB}{225,237,237}
\caption{Intra-judge stability on MultiModal (30 tasks). Three independent runs with the same GPT judge configuration. Each dimension reports scores from Run 1 / Run 2 / Run 3, followed by the mean and standard deviation.}
\label{tab:intra-judge}
\renewcommand\arraystretch{1.15}
\setlength{\tabcolsep}{3pt}
\begin{tabularx}{\linewidth}{@{} l | Y Y Y | Y Y Y | Y Y Y | Y Y Y | Y Y @{}}
\toprule
\multirow{2}{*}{\textbf{Model}} & \multicolumn{3}{c|}{\textbf{Synthesis}} & \multicolumn{3}{c|}{\textbf{Factuality}} & \multicolumn{3}{c|}{\textbf{Process}} & \multicolumn{3}{c|}{\textbf{Overall}} & \multirow{2}{*}{\textbf{Avg}} & \multirow{2}{*}{\textbf{Std}} \\
\cmidrule(lr){2-4} \cmidrule(lr){5-7} \cmidrule(lr){8-10} \cmidrule(lr){11-13}
& R1 & R2 & R3 & R1 & R2 & R3 & R1 & R2 & R3 & R1 & R2 & R3 & & \\
\midrule
OpenAI Deep Research & 66.7 & 66.5 & 66.7 & 77.0 & 75.0 & 76.9 & 66.8 & 65.0 & 66.2 & 70.2 & 68.8 & 69.9 & 69.6 & 0.6 \\
Gemini-3.1-Pro & 66.4 & 66.3 & 66.9 & 73.7 & 70.0 & 72.7 & 64.1 & 63.0 & 62.8 & 68.1 & 66.4 & 67.5 & 67.3 & 0.6 \\
\midrule
\rowcolor{rowhl} MiroThinker-1.7 & 69.0 & 68.7 & 68.6 & 78.5 & 80.9 & 79.6 & 67.4 & 67.7 & 67.4 & 71.6 & 72.4 & 71.9 & 72.0 & 0.3 \\
\rowcolor{rowhl} MiroThinker-H1 & 71.5 & 72.0 & 70.8 & 78.4 & 76.2 & 77.4 & 73.5 & 73.6 & 73.3 & 74.5 & 73.9 & 73.8 & 74.1 & 0.3 \\
\bottomrule
\end{tabularx}
\end{table*}

The standard deviations on Overall are remarkably low, ranging from 0.3 (MiroThinker-H1 and MiroThinker-1.7) to 0.6 (OpenAI Deep Research and Gemini-3.1-Pro), and the system ranking is identical across all three runs. At the sub-dimension level, Synthesis scores are the most stable with variations under 1 point, while Factuality shows slightly larger fluctuations of up to 3 points for individual systems (e.g., Gemini-3.1-Pro: 73.7 / 70.0 / 72.7), likely due to the stochastic nature of web search during claim verification. Despite these per-dimension fluctuations, the Overall ranking remains perfectly preserved, confirming that the evaluation results are stable under repeated execution.

\subsection{Cross-Judge Consistency and Prompt Sensitivity}
\label{app:prompt-sensitivity}

Beyond run-level variance, we further examine whether system rankings are robust to the choice of judge model and the formulation of judge prompts. Table~\ref{tab:judge-robustness} summarizes both experiments. Each cell reports scores in the format \textit{original / alternative / $\Delta$}.

\paragraph{Cross-Judge Consistency.}
We re-evaluate the Text-Only setting (70 tasks) using Gemini as an alternative judge for all three dimensions, covering six systems (Gemini-2.5-Pro for synthesis and process evaluation, and Gemini-3-Flash for factuality evaluation). The Gemini judge produces substantially higher absolute scores across the board, with Overall deltas ranging from +13.2 (OpenAI Deep Research and MiroThinker-H1) to +16.9 (ChatGLM Agent). This systematic inflation is most pronounced on Process (deltas of +16.6 to +21.3) and least on Factuality (+6.0 to +11.7), suggesting that Gemini applies more lenient criteria for process evaluation than for factual verification. Crucially, despite these large absolute shifts, the relative ranking of all six systems is perfectly preserved ($\Delta$Rank = 0 for every system), yielding a Kendall's $\tau$ of 1.0 on Overall. This demonstrates that cross-judge differences are systematic rather than selective, affecting all systems similarly and leaving comparative conclusions intact.

\paragraph{Prompt Sensitivity.}
We re-evaluate four systems on the MultiModal setting (30 tasks) using the same GPT judge but with a modified prompt that rephrases the scoring criteria in a more concise format and adjusts the ordering of evaluation dimensions. In contrast to the cross-judge experiment, the prompt modification produces only minimal score changes: Overall deltas range from $-$0.5 (MiroThinker-H1) to $-$1.6 (OpenAI Deep Research), with most per-dimension shifts below 1 point. The only dimension showing slightly larger variation is Factuality (up to $-$2.7 for Gemini-3.1-Pro), consistent with the higher sensitivity of claim-level verification to prompt phrasing. As with the cross-judge experiment, system rankings are fully preserved ($\Delta$Rank = 0), confirming that the evaluation outcomes are robust to reasonable prompt reformulations.

\begin{table*}[t]
\centering
\footnotesize
\definecolor{rowhl}{RGB}{225,237,237}
\caption{Robustness to judge model choice and prompt variation. Each cell reports \textit{original\,/\,alternative\,/\,$\Delta$}. $\Delta$Rank: rank change based on Overall score. \textbf{Upper}: Cross-judge consistency on Text-Only (70 tasks), GPT vs.\ Gemini. \textbf{Lower}: Prompt sensitivity on MultiModal (30 tasks), original vs.\ modified prompt with the same GPT judge.}
\label{tab:judge-robustness}
\renewcommand\arraystretch{1.15}
\setlength{\tabcolsep}{4pt}
\begin{tabularx}{\linewidth}{@{} l | Y Y Y Y | c @{}}
\toprule
\textbf{Model} & \textbf{Synthesis} & \textbf{Factuality} & \textbf{Process} & \textbf{Overall} & $\Delta$\textbf{Rank} \\
\midrule
\multicolumn{6}{c}{\textit{Cross-Judge: Text-Only (70 Tasks) --- GPT Series (orig.) / Gemini Series (alt.) / $\Delta$}} \\
\midrule
OpenAI Deep Research & 73.8\,/\,90.2\,/\,+16.4 & 83.3\,/\,89.3\,/\,+6.0 & 73.1\,/\,90.2\,/\,+17.1 & 76.7\,/\,89.9\,/\,+13.2 & 0 \\
Gemini-3.1-Pro & 71.2\,/\,89.5\,/\,+18.3 & 71.3\,/\,81.8\,/\,+10.5 & 67.1\,/\,87.9\,/\,+20.9 & 69.9\,/\,86.4\,/\,+16.5 & 0 \\
ChatGLM Agent & 63.2\,/\,82.2\,/\,+19.0 & 68.6\,/\,80.3\,/\,+11.7 & 65.6\,/\,85.7\,/\,+20.1 & 65.8\,/\,82.7\,/\,+16.9 & 0 \\
\rowcolor{rowhl} MiroThinker-1.7-mini & 74.0\,/\,90.3\,/\,+16.3 & 76.2\,/\,86.2\,/\,+10.0 & 68.5\,/\,89.8\,/\,+21.3 & 72.9\,/\,88.8\,/\,+15.9 & 0 \\
\rowcolor{rowhl} MiroThinker-1.7 & 74.3\,/\,90.8\,/\,+16.5 & 79.4\,/\,87.6\,/\,+8.2 & 72.7\,/\,91.0\,/\,+18.3 & 75.5\,/\,89.8\,/\,+14.3 & 0 \\
\rowcolor{rowhl} MiroThinker-H1 & 76.7\,/\,92.1\,/\,+15.4 & 81.1\,/\,88.6\,/\,+7.5 & 74.7\,/\,91.3\,/\,+16.6 & 77.5\,/\,90.7\,/\,+13.2 & 0 \\
\midrule
\multicolumn{6}{c}{\textit{Prompt Sensitivity: MultiModal (30 Tasks) --- Original / Modified / $\Delta$}} \\
\midrule
OpenAI Deep Research & 66.7\,/\,66.3\,/\,-0.4 & 77.0\,/\,74.6\,/\,-2.4 & 66.8\,/\,65.0\,/\,-1.8 & 70.2\,/\,68.6\,/\,-1.6 & 0 \\
Gemini-3.1-Pro & 66.4\,/\,66.3\,/\,-0.1 & 73.7\,/\,71.0\,/\,-2.7 & 64.1\,/\,63.0\,/\,-1.0 & 68.1\,/\,66.8\,/\,-1.3 & 0 \\
\rowcolor{rowhl} MiroThinker-1.7 & 69.0\,/\,68.8\,/\,-0.2 & 78.4\,/\,76.3\,/\,-2.1 & 67.4\,/\,67.5\,/\,+0.2 & 71.6\,/\,70.9\,/\,-0.7 & 0 \\
\rowcolor{rowhl} MiroThinker-H1 & 71.5\,/\,71.7\,/\,+0.2 & 78.5\,/\,77.9\,/\,-0.6 & 73.5\,/\,72.5\,/\,-1.0 & 74.5\,/\,74.0\,/\,-0.5 & 0 \\
\bottomrule
\end{tabularx}
\end{table*}

\subsection{Human Study}
\label{app:human-study}

To validate that our automated evaluation aligns with expert human judgment, we conduct a human study with 5 volunteers. We randomly sample 5 queries from the benchmark and collect reports from all deep research systems that support multimodal attachments. For each case, annotators are provided with both the final report and the associated research process, and are asked to rank the systems based on overall quality, jointly considering the effectiveness of the research process and the quality of the resulting report.

Table~\ref{tab:human-ranking} reports the average human ranking alongside the MiroEval ranking for each system. The two rankings exhibit strong agreement, with Kendall's $\tau$ = 0.91 and Spearman's $\rho$ = 0.95. The top three systems under human judgment (MiroThinker-H1, OpenAI Deep Research, MiroThinker-1.7) match the top three under MiroEval exactly, and the largest rank shift across all systems is only 2 positions (Qwen-3.5-Plus).

\begin{table}[t]
\centering
\footnotesize
\caption{Comparison between human rankings and MiroEval rankings. Human rankings are averaged across 5 annotators. $\Delta$Rank: positive values indicate higher human ranking than MiroEval.}
\label{tab:human-ranking}
\renewcommand\arraystretch{1.1}
\setlength{\tabcolsep}{4pt}
\begin{tabular}{lccc}
\toprule
\textbf{System} & \textbf{Human} & \textbf{MiroEval} & $\Delta$\textbf{Rank} \\
\midrule
MiroThinker-H1 & 1.8 & 1 & $=$ \\
OpenAI Deep Research & 2.5 & 2 & $=$ \\
MiroThinker-1.7 & 2.8 & 3 & $=$ \\
Claude-Opus-4.6 Research & 5.2 & 5 & $\uparrow 1$ \\
Gemini-3.1-Pro Deep Research & 5.3 & 4 & $\downarrow 1$ \\
MiniMax-M2.5 Research & 6.0 & 6 & $=$ \\
Qwen-3.5-Plus Deep Research & 6.8 & 9 & $\uparrow 2$ \\
ChatGLM Agent & 7.3 & 7 & $\downarrow 1$ \\
Manus-1.6-Max Wide Research & 8.0 & 8 & $\downarrow 1$ \\
Grok Deep Research & 9.5 & 10 & $=$ \\
\bottomrule
\end{tabular}
\end{table}

\paragraph{Summary.}
Across all four analyses, the relative ranking of systems is remarkably stable. Repeated runs produce Overall standard deviations below 0.6; switching from GPT to Gemini as judge inflates absolute scores by 13 to 17 points but preserves the ranking perfectly; modifying the judge prompt shifts scores by less than 2 points with no rank changes; and expert human annotators converge on the same top-tier systems as MiroEval with a maximum rank shift of 2 positions. These results confirm that the comparative conclusions drawn in the main text are robust to the evaluation configuration, and that absolute score differences between judge models reflect systematic calibration offsets rather than meaningful disagreements about relative system quality.

\definecolor{rowhl}{RGB}{225,237,237}

\newtcolorbox{casebox}[1]{
    breakable,
    colback=rowhl!20,            
    colframe=rowhl!80!black,     
    colbacktitle=rowhl!80!black, 
    coltitle=white,            
    fonttitle=\bfseries\large,
    title=#1,
    boxrule=0.6pt,
    arc=3pt,
    left=8pt,
    right=8pt,
    top=6pt,
    bottom=6pt
}

\clearpage
\section{Case Study}

\subsection{Synthesis Evaluation}

We present two representative case studies to illustrate how the adaptive synthesis quality evaluation operates on concrete tasks, including the generated dimensions, criteria, and scoring. Both cases are multimodal tasks with attachment-augmented queries.

\begin{casebox}{Case 1: Analyzing the 50 Fastest-Growing Software Vendors}

\textbf{Task.}
The user provides a screenshot of the BREX BENCHMARK ``50 fastest-growing software vendors of 2025'' ranking (an image listing vendor names and ranks from \#01~Cursor to \#50~Wispr) and asks the system to extract all company names, research each company's product, business model, culture, and competitive advantages, and produce an integrated report.

\medskip
\textbf{Extracted Key Facts.}
The framework extracts the following factual anchors from the attachment:
\begin{itemize}[nosep,leftmargin=*]
\item Top-10 roster: 01~Cursor, 02~OpenRouter, 03~Kling~AI, 04~Retell~AI, 05~Perplexity, 06~Windsurf, 07~FireCrawl, 08~Clay, 09~Replit, 10~Exa.
\item The graphic contains \emph{only} vendor names and ranks; no growth rates, categories, business models, or financial metrics are shown.
\item Multiple vendors include ``.ai'' in their names, but this alone does not establish AI-native status.
\end{itemize}

\medskip
\textbf{Generated Evaluation Dimensions and Weights.}
Beyond the four fixed dimensions, three task-specific Grounding \& Expertise dimensions are generated. The assigned weights are:

\smallskip
\begin{tabular}{lc}
\toprule
\textbf{Dimension} & \textbf{Weight} \\
\midrule
Coverage & 0.27 \\
Insight & 0.30 \\
Instruction-following & 0.14 \\
Clarity & 0.09 \\
\rowcolor{rowhl} Attachment-Grounded Scope \& Methodology Integrity & 0.08 \\
\rowcolor{rowhl} Evidence-Linked Business Model \& AI-Native Typology & 0.07 \\
\rowcolor{rowhl} Investment-Grade Profitability \& Go-to-Market Evaluation & 0.05 \\
\bottomrule
\end{tabular}

\smallskip
The first grounding dimension checks whether the report uses the exact top-20 roster as its sampling frame and avoids inferring attributes from the graphic. The second checks whether business-model labels are backed by external evidence rather than assumed. The third checks whether profitability claims are sourced from credible references.

\medskip
\textbf{Cross-System Scoring.}
We compare MiroThinker-H1 (overall: 8.5) against ChatGLM Agent (overall: 4.5).

\smallskip
\emph{Coverage --- ``Complete per-company classifications for the top-20.''}

MiroThinker-H1 provides a per-company table classifying all 20 vendors by business model, customer segment, and technology route. The evaluator notes: ``\textit{The report provides a table that lists all top-20 companies with columns for business model, core customer group, and technology route, which it uses to indicate AI-native status.}'' Score: \textbf{9.4}.

ChatGLM Agent does not produce a per-company classification table. The evaluator notes: ``\textit{The report does not provide a per-company matrix or table that classifies each of the top-20 by business model, primary customer segment, and AI-native vs.\ traditional. It relies on anecdotes and placeholders.}'' Score: \textbf{1.0}.

\smallskip
\emph{Grounding: Evidence-Linked Business Model \& Typology --- ``Correct top-20 sampling frame and coverage.''}

MiroThinker-H1 explicitly lists all 20 companies with correct names and ranks matching the attachment. The evaluator notes: ``\textit{The report clearly works off the correct Brex `top 20' roster. It explicitly lists all 20 companies in Table~1 with correct names and ranks matching the key facts.}'' Score: \textbf{9.8}.

ChatGLM Agent introduces companies outside the top-20 and conflates them into its analysis. The evaluator notes: ``The report repeatedly goes beyond the top-20 scope, using ElevenLabs---not in the roster---as a core example for business model and profitability conclusions.'' Score: \textbf{2.5}.

\smallskip
\emph{Grounding: Scope \& Methodology Integrity --- ``Acknowledges graphic content limitations and avoids inference.''}

ChatGLM Agent fabricates growth rates from the graphic (e.g., ``Cursor 1000\%, Kling AI 1900\%''), when the attachment contains no growth data at all. The evaluator notes: ``\textit{The attachment shows only names and ranks. The report neither acknowledges this limitation nor avoids inference; it asserts growth rates, profitability, and ARR figures not present in the graphic.}'' Score: \textbf{0.5}.

\medskip
\textbf{Takeaway.}
The dynamically generated grounding dimension catches a critical failure mode---fabricating growth rates and financials from a graphic that contains only names and ranks---that a fixed rubric would not detect. The key-facts extraction step transforms the abstract instruction into precise checkpoints (e.g., ``does the report use the exact top-20 roster?''), enabling fine-grained discrimination.

\end{casebox}

\begin{casebox}{Case 2: Veterinary Nutrition Planning for a Senior Cat}

\textbf{Task.}
A user provides medical records and 10 product photographs (canned wet foods from brands including Schesir After Dark, RAWZ, AIXIA, Unicharm, Zealandia, and Miaw Miaw) for a 12-year-old cat with chronic pancreatitis and mild CKD. The system must compare the nutritional profiles and produce a prioritized feeding recommendation.

\medskip
\textbf{Extracted Key Facts.}
The framework extracts product-level facts from each photograph. Crucially, most product images do \emph{not} display nutritional analysis panels:

\begin{itemize}[nosep,leftmargin=*]
\item AIXIA Kenko-can 11+: phosphorus 0.08\%, sodium 0.10\% (visible on label); \textbf{no fat content shown}.
\item Schesir After Dark Chicken 80\,g: labeled ``Complete,'' ``Grain Free''; \textbf{no fat or phosphorus visible}.
\item RAWZ Shredded Chicken \& Pumpkin $\sim$3\,oz: \textbf{no nutrient panel visible}.
\item Unicharm Silver Spoon 13+ (multiple tuna variants, 70\,g each): \textbf{no nutrient values visible}.
\end{itemize}

\medskip
\textbf{Generated Evaluation Dimensions and Weights.}
Three task-specific dimensions are generated:

\smallskip
\begin{tabular}{lc}
\toprule
\textbf{Dimension} & \textbf{Weight} \\
\midrule
Coverage & 0.28 \\
Insight & 0.26 \\
Instruction-following & 0.16 \\
Clarity & 0.10 \\
\rowcolor{rowhl} SKU-Verified Nutrient Provenance \& Regional Fidelity & 0.07 \\
\rowcolor{rowhl} Clinically-Normalized Nutrition Modeling \& Suitability Ranking & 0.09 \\
\rowcolor{rowhl} Uncertainty-Aware Data Gap Governance & 0.04 \\
\bottomrule
\end{tabular}

\smallskip
The first grounding dimension checks whether nutrient values are matched to the exact SKU and regional variant (e.g., Japan-market vs.\ US-market formulations). The second checks whether nutrients are converted to comparable clinical metrics for pancreatitis and CKD. The third checks whether the report explicitly acknowledges missing data.

\medskip
\textbf{Cross-System Scoring.}
We compare Gemini-3.1-Pro (overall: 7.6) against MiniMax-M2.5 (overall: 4.0).

\smallskip
\emph{Insight --- ``Trade-off reasoning and decision logic.''}

Gemini-3.1-Pro constructs a two-dimensional fat--phosphorus framework and proposes mixed-feeding strategies. The evaluator notes: ``\textit{Demonstrates nuanced balancing with a 2D fat--phosphorus framework and proposes tactical mixed feeding to reconcile low-P vs low-fat trade-offs. It classifies products by which axis they satisfy.}'' Score: \textbf{8.9}.

MiniMax-M2.5 assigns subjective star ratings without explicit trade-off analysis. The evaluator notes: ``\textit{Trade-offs are only superficially addressed. The report assigns star ratings but does not explicitly weigh key compromises.}'' Score: \textbf{4.6}.

\smallskip
\emph{Grounding: Uncertainty-Aware Data Gap Governance --- ``Explicitly acknowledges missing fat and phosphorus for most SKUs and avoids inventing values.''}

Gemini-3.1-Pro partially acknowledges missing panels and marks some values as unknown, but still uses retail aggregator data without confirming regional SKU matches. Score: \textbf{3.2}.

MiniMax-M2.5 asserts specific nutrient values (e.g., ``Silver Spoon fat: 0.3\%,'' ``RAWZ fat: 1.5\%'') for products whose labels show \emph{no nutritional data at all}, without citing any source. The evaluator notes: ``\textit{Attachments show no fat/phosphorus panels for Schesir After Dark, RAWZ, Unicharm Silver Spoon 13+, SEEDS Golden Cat, Zealandia Wallaby. The report nonetheless provides specific values without acknowledging data gaps or citing external sources.}'' Score: \textbf{0.8}.

\smallskip
\emph{Grounding: SKU-Verified Nutrient Provenance --- ``Authoritative, region-matched nutrient sourcing.''}

Both systems struggle with regional fidelity. Gemini-3.1-Pro sources from retail aggregators rather than manufacturer datasheets, and sometimes conflates Japan-market and US-market variants. Score: \textbf{3.4}. MiniMax-M2.5 provides no manufacturer URLs or datasheets for any value. Score: \textbf{0.5}.

\medskip
\textbf{Takeaway.}
The \emph{Uncertainty-Aware Data Gap Governance} dimension---generated because the key-facts extraction identified that most product images lack nutrient panels---catches a failure mode where systems fabricate precise-looking numbers from non-existent label data. A fixed rubric evaluating only ``nutritional accuracy'' would miss this distinction, as it cannot know \emph{a priori} which data points are available in the attachments.

\end{casebox}

\paragraph{Overall observation.}
The two cases illustrate how adaptive evaluation captures failure modes invisible to fixed rubrics. In Case~1, the key-facts extraction reveals that the attachment contains only names and ranks, prompting a grounding dimension that penalizes fabricated growth rates. In Case~2, the extraction identifies pervasive missing nutrient panels, prompting an uncertainty-governance dimension that penalizes invented values. In both cases, the dynamically generated criteria provide task-specific discrimination, while the shared fixed dimensions maintain cross-task comparability.

\clearpage
\subsection{Factuality Evaluation}

We present several representative case studies to illustrate typical examples of verifying a statement using the proposed agentic evaluation framework. We also include two cases where the judgments are incorrect; however, it should be noted that such cases only occur occasionally.

\begin{casebox}{Correct Case 1}

\textbf{Extracted Statement}

In the second half of 2014, global oversupply driven by U.S.\ shale production growth and slowing demand caused oil prices to plummet from approximately \$110 to below \$40 per barrel.

\medskip
\textbf{Agentic Verification:} Wrong.

\textbf{Reasoning:}
While the qualitative explanation---U.S.\ shale-driven oversupply combined with weakening demand---is well supported by EIA and World Bank analyses, the specific numeric claim that prices fell ``below \$40'' during H2 2014 is incorrect. EIA monthly data show Brent crude averaging approximately \$106.77 in July 2014 and \$62.34 in December 2014. WTI ended December 2014 at approximately \$59.29. Prices did not fall below \$40 per barrel until early 2016, more than a year after the period referenced in the statement.

\medskip
\textbf{Evidence}
\begin{itemize}
\item \textit{EIA (Brent)}: ``EIA monthly table for 2014: Brent July 2014
$\approx$ \$106.77; Brent December 2014 $\approx$ \$62.34.''
\item \textit{EIA (WTI)}: ``WTI December 2014 $\approx$ \$59.29 (not below
\$40 at year-end 2014).''
\end{itemize}

\medskip
\textbf{Human Annotation} 

Right

\medskip
\textbf{Final Judge}

The model judgment is correct. Although the qualitative explanation is accurate, the claim that prices fell ``below \$40'' in H2 2014 is false. EIA data show end-2014 prices near \$62 (Brent) and \$59 (WTI), with sub-\$40 levels only reached in early 2016. This case illustrates a common hallucination pattern: correct directional reasoning but incorrect magnitude or timing.

\end{casebox}

\begin{casebox}{Correct Case 2}

\textbf{Statement}

In 2019, Huawei reported revenue of CNY 858,833M, operating profit of CNY 77,835M, and an operating margin of 9.1\%.

\medskip
\textbf{Agentic Verification:} Right.

\textbf{Reasoning:}
The system extracted all three figures from the five-year financial highlights table in the user-uploaded screenshot, and then independently retrieved the same table from Huawei's 2019 Annual Report. All three values---revenue, operating profit, and the stated margin---are consistent across both sources. An internal arithmetic cross-check further confirms the margin: $77{,}835 \div 858{,}833 \approx 9.07\%$, which rounds to the reported $9.1\%$.

\medskip
\textbf{Evidence}
\begin{itemize}
\item \textit{Uploaded screenshot (image)}: ``Revenue row: \ldots 858,833~\ldots; Operating profit row: \ldots 77,835~\ldots; Operating margin row: \ldots 9.1\%~\ldots''
\item \textit{Huawei 2019 Annual Report (PDF)}: ``Five-Year Financial Highlights (2019 column): Revenue (CNY Million) 858,833; Operating profit 77,835; Operating margin 9.1\%.''
\end{itemize}

\medskip
\textbf{Human Annotation}

Right

\medskip
\textbf{Final Judge}

Both the model and the human annotator correctly judge the statement as accurate. The three figures are directly verifiable from Huawei's publicly available 2019 Annual Report, and the margin is arithmetically consistent with the revenue and profit values: $77,835 \div 858,833 = 9.07\% \approx 9.1\%$. This case illustrates a distinctive multimodal verification workflow: the system first extracts numeric data from an embedded image (a screenshot of the financial highlights table), then independently confirms the same figures via an authoritative external document, and finally validates internal arithmetic consistency---a three-layer cross-check that is not possible in text-only settings.

\end{casebox}

\begin{casebox}{Correct Case 3}

\textbf{Statement}

The LiuGong 856HE electric wheel loader is equipped with a CATL LFP battery rated at 350~kWh / 580~V, with a charging time of under 75~minutes.

\medskip
\textbf{Agentic Verification:} Conflict.

\textbf{Reasoning:}
The attached technical review explicitly reports a battery capacity of \textbf{350~kWh} and a rated voltage of \textbf{580~V}, with a charging time under \textbf{75~minutes}. However, LiuGong's official product page lists the 856HE with an LFP battery rated at \textbf{432~kWh}, resulting in an 82~kWh (23\%) discrepancy. Since both sources are credible yet inconsistent, and the difference cannot be explained by rounding, the claim cannot be definitively verified without further clarification (e.g., model variants or specification updates).

\medskip
\textbf{Evidence}
\begin{itemize}
\item \textit{Attachment --- technical review PDF (\emph{Electric Wheel Loader Overview.pdf}, p.~8)}: ``the LiuGong 856HE EWL \ldots\ has a rated energy capacity of \textbf{350~kWh} and a rated voltage of 580~V, with a charging time of under 75~min.''
\item \textit{Web --- LiuGong North America product page (856HE)}: ``Battery Type: LFP. Rated Storage Energy: \textbf{432~kWh}. Operating Weight: 46,297~lb (21,000~kg).''
\end{itemize}

\medskip
\textbf{Human Annotation}

Conflict

\medskip
\textbf{Final Judge}

The model judgment is correct. The discrepancy between 350~kWh and 432~kWh indicates conflicting specifications across credible sources, likely due to different model variants or updates. This case highlights a common multimodal conflict scenario: consistent evidence within each source but irreconcilable differences across sources.

\end{casebox}

\begin{casebox}{Incorrect Case 1}

\textbf{Statement}

\textit{Star Raiders} was released in 1979.

\medskip
\textbf{Agentic Verification:} Wrong.

\textbf{Reasoning:}
The system retrieved the Wikipedia article for \textit{Star Raiders} and identified the phrase ``published in 1980 by Atari, Inc.'', concluding that the correct release year is 1980 and that 1979 likely refers only to development or internal demonstrations.

\medskip
\textbf{Evidence}
\begin{itemize}
  \item \textit{Wikipedia --- Star Raiders}: ``Star Raiders is a space combat simulator video game created by Doug Neubauer and published in 1980 by Atari, Inc. Originally released for the Atari 400/800 computers\ldots'' [excerpt truncated]
\end{itemize}

\medskip
\textbf{Human Annotation}

Right

\medskip
\textbf{Final Judge}

The agentic verdict is incorrect. While the retrieved evidence emphasizes the 1980 publication date, \textit{Star Raiders} was already available in 1979 as a pack-in title for the Atari 400/800 launch. Therefore, the statement is factually valid under a broader definition of release. This case highlights a temporal ambiguity: different sources refer to different release events, and the agent fails to reconcile these interpretations.

\end{casebox}

\begin{casebox}{Incorrect Case 2}

\textbf{Statement}

Revenue from products built on generative AI models grew over 200\% year-over-year.

\medskip
\textbf{Agentic Verification}

\textbf{Result:} Right.

\textbf{Reasoning:}
The system retrieved the CEO's remarks from both the uploaded earnings document and the company's official blog, which report the figure as ``nearly 400\% year-over-year.'' Since 400\% exceeds 200\%, the system interpreted the statement ``over 200\%'' as a conservative but numerically valid description and returned a Right verdict.

\medskip
\textbf{Evidence}
\begin{itemize}
  \item \textit{Attachment --- Alphabet Q4 2025 Earnings Release (PDF, p.\,1)}: ``In Q4, revenue from products built on our generative AI models grew \textbf{nearly 400\%} year-over-year.''
  \item \textit{Web --- Alphabet CEO Blog}: ``nearly 400\% year-over-year.'' (\url{https://blog.google/company-news/inside-google/message-ceo/alphabet-earnings-q4-2025/})
\end{itemize}

\medskip
\textbf{Human Annotation}

Wrong

\medskip
\textbf{Final Judge}

The model judgment is incorrect. Although ``nearly 400\%'' logically satisfies the lower-bound claim ``over 200\%,'' the statement materially understates the reported magnitude and thus misrepresents the underlying fact. This case highlights a common failure mode: treating logical entailment as sufficient for factual correctness, particularly for vague lower-bound expressions (e.g., ``over X\%''), leading to false positives.

\end{casebox}

\clearpage
\subsection{Process Evaluation}

\paragraph{Process Case Studies.}
To make the process evaluation more interpretable, we present two representative case studies: we first show how raw trajectories are abstracted into structured process representations, and then compare systems based on both intrinsic process quality and process--report alignment. The first case highlights an evidence-sparse text-only task, where strong performance depends on scope control and conservative synthesis. The second highlights a multimodal task, where the key difference is whether the attachment is incorporated as an early grounding constraint that shapes the subsequent investigation.

\begin{casebox}{Case Box 1: Text-only task with fragmented and incomplete evidence}
\textbf{Task.}
The task asks the system to analyze the employment destinations of graduates from China’s 985 universities across 2023--2025, identify the major recurring employers, compare hiring scale and industry absorption, distinguish undergraduate from postgraduate outcomes, and exclude public-sector placements.

\textbf{Structured process representation.}
After converting raw trajectories into atomic process units, the three systems exhibit markedly different structures. \textit{MiroThinker-H1} follows a compact trajectory of \textbf{plan $\rightarrow$ search $\rightarrow$ read $\rightarrow$ search $\rightarrow$ read $\rightarrow$ analyze $\rightarrow$ verify $\rightarrow$ synthesize}. Its process is anchored by an early scope decision: only \emph{enterprise employment} is counted, while government agencies, public institutions, further study, and entrepreneurship are explicitly excluded. This decision determines all subsequent evidence collection. The extracted process findings show that H1 identifies three core constraints: (1) public evidence is structurally incomplete and cannot support a fully auditable cross-985 ranking; (2) many university reports mix enterprise and non-enterprise destinations, so employer rankings require unit-type cleaning; and (3) company-side recruiting plans can serve as auxiliary scale signals but are not equivalent to realized graduate placements.

By contrast, \textit{MiroThinker-v17} shows a much longer but less consolidated structure. Although it retrieves more local evidence and successfully gathers several useful university-level records, its trajectory is dominated by repeated search and scrape actions, with relatively weak transition into explicit verification and final synthesis. \textit{MiroThinker} is even more search-heavy: its process remains largely at the level of searching for candidate sources and identifying possible clues, but never reaches a stable stage of cross-source consolidation.

\textbf{Cross-model comparison.}
This structural difference explains the process scores. H1 does not achieve the widest search breadth, but it performs better on efficiency and alignment because it converts retrieved evidence into a bounded conclusion. In particular, it explicitly states that the available evidence supports only a \emph{candidate set of major employers}, not a reliable Top-10 ranking. This yields the best alignment profile in the case: the report remains largely traceable to the process, and unsupported extrapolations are limited. In contrast, both v17 and MiroThinker introduce more conclusions that go beyond what was actually established in the trajectory. Their weakness is therefore not simply insufficient search, but insufficiently grounded synthesis.

\textbf{Takeaway.}
This case illustrates a central failure mode in evidence-sparse research tasks: stronger processes are not necessarily those that collect more raw material, but those that define the scope early, verify the comparability of evidence, and stop at the evidence boundary instead of fabricating a complete answer.
\end{casebox}

\begin{casebox}{Case Box 2: Multimodal task where attachment grounding changes the trajectory}
\textbf{Task.}
The task asks the system to assess the climate effects of a hypothetical geographic intervention in which the Tarim Basin becomes an inland sea connected to the Indian Ocean, using both historical analogues and quantitative climate evidence from attached materials and external sources.

\textbf{Structured process representation.}
The multimodal case reveals a different kind of process difference. \textit{MiroThinker-H1} reads the attachment at the very beginning and extracts a key physical constraint: the new waterway crosses the Pamir/Karakoram region. This attachment-grounded constraint becomes the starting point for the later investigation. After structuring the trajectory, H1 follows a layered process of \textbf{read attachment $\rightarrow$ retrieve analogues $\rightarrow$ read climate studies $\rightarrow$ analyze mechanisms $\rightarrow$ synthesize}. Its extracted findings form an ``evidence pyramid'': modern inland-sea analogues, paleoclimate analogies, reverse evidence from sea retreat and aridification, and circulation-level explanations from climate-model studies. Its Search:Read ratio is also highly distinctive, indicating that once the direction is established, the model spends most of its budget digesting evidence rather than repeatedly re-searching.

\textit{MiroThinker-v17} retrieves a larger number of quantitative fragments and surfaces several useful numerical observations, but its trajectory is less tightly organized around a single evidential hierarchy. \textit{MiroThinker} shows the weakest structured process: it relies more heavily on broad analogy search and limited reading, with several final numerical ranges functioning as inferential placeholders rather than findings directly supported by the process.

\textbf{Cross-model comparison.}
These structural differences are reflected most clearly in alignment. H1 achieves the best process--report consistency because most major report claims can be linked back to process-level findings, and the report contains few substantive contradictions. By contrast, v17 introduces some unsupported extrapolations beyond the trace, while MiroThinker includes multiple quantitative ranges that are not directly grounded in the retrieved evidence. The core distinction is therefore not simply whether a model can retrieve climate analogies, but whether it uses the attachment to define the scenario constraints early and preserves that grounding throughout the report.

\textbf{Takeaway.}
This case shows why multimodal process evaluation cannot be reduced to output inspection alone. A model may produce a plausible report, yet still fail to ground its reasoning in the attachment that defines the task. Strong multimodal research processes are characterized by early attachment integration, evidence digestion over repeated search, and tighter traceability between intermediate findings and final conclusions.
\end{casebox}

\paragraph{Overall observation.}
Together, these two cases illustrate the value of process-centric evaluation beyond final-report scoring. In the text-only case, the decisive factor is disciplined scope control under fragmented evidence; in the multimodal case, it is whether the attachment becomes a first-class constraint in the research trajectory. Across both settings, the strongest processes share the same procedural pattern: early task reframing, selective evidence digestion, explicit handling of limitations or conflicts, and conservative synthesis that stays within the support of the documented trajectory.

\end{document}